\documentclass[10pt,journal,compsoc]{IEEEtran}
\ifCLASSOPTIONcompsoc
  \usepackage[nocompress]{cite}
\else
  \usepackage{cite}
\fi
\ifCLASSINFOpdf
  \usepackage[pdftex]{graphicx}
    \graphicspath{ {./figures/} }
    \usepackage[outdir=./]{epstopdf}
\else
  \usepackage[dvips]{graphicx}
\fi
\usepackage{booktabs}
\usepackage{enumitem}
\usepackage{caption}

\usepackage{amsmath}
\usepackage{amssymb}

\usepackage[utf8]{inputenc}
\usepackage[english]{babel}

\usepackage{tabu}
\usepackage[ruled]{algorithm2e} 
\usepackage{algorithmic}

\usepackage{array}

\ifCLASSOPTIONcompsoc
 \usepackage[caption=false,font=footnotesize,labelfont=sf,textfont=sf]{subfig}
\else
 \usepackage[caption=false,font=footnotesize]{subfig}
\fi

\makeatletter
\renewcommand{\fnum@figure}{Fig. \thefigure}
\makeatother

\begin{document}
%
\title{Reputation Bootstrapping for Composite Services using CP-nets}

\author{Sajib Mistry and
        Athman Bouguettaya, \IEEEmembership{Fellow,~IEEE} 
\IEEEcompsocitemizethanks{\IEEEcompsocthanksitem S. Mistry is with School of Electrical Engineering, Computing and Mathematical Sciences, Curtin University, and A. Bouguettaya is with the School of Computer Science, University of Sydney, Australia.
\protect \\ E-mail: sajib.mistry@curtin.edu.au, athman.bouguettaya@sydney.edu.au

}
\thanks{}}

\IEEEtitleabstractindextext{%
\begin{abstract}
We propose a novel framework to bootstrap the reputation of on-demand service compositions. On-demand compositions are usually context-aware and have little or no direct consumer feedback. The reputation bootstrapping of single or atomic services do not consider the topology of the composition and relationships among reputation-related factors. We apply Conditional Preference Networks (CP-nets) of reputation-related factors for each of component services in a composition. The reputation of a composite service is bootstrapped by the composition of CP-nets. We consider the history of invocation among component services to determine reputation-interdependence in a composition. The composition rules are constructed using the composition topology and four types of reputation-influence among component services. A heuristic-based Q-learning approach is proposed to select the optimal set of reputation-related CP-nets. Experimental results prove the efficiency of the proposed approach.
\end{abstract}

\begin{IEEEkeywords}
Reputation Bootstrapping, Composite Services, CP-nets, Reputation propagation heuristics,  Combinatorial selection. \vspace{-2mm}
\end{IEEEkeywords}}

\maketitle

\IEEEdisplaynontitleabstractindextext

%
\IEEEpeerreviewmaketitle

\vspace{-20mm}
\IEEEraisesectionheading{\section{Introduction}\label{sec:introduction}}

%
%
%
%

\IEEEPARstart{R}{eputation} is an important Quality of Service (QoS) in a service-oriented environment \cite{Erl:2004:SAF:983556,DBLP:journals/internet/MalikB09,DBLP:conf/icws/BahutairBN20}. Valid consumers' \textit{direct feedback} or \textit{ratings} are often used as a reputation indicator \cite{DBLP:conf/trustcom/JiaoLLL11,6461455} to select the reputable services. However, direct consumer feedback is not usually available for new on-demand compositions \cite{4279708}. An on-demand composition request usually results in a new topology by adding newly published component services or removing some of the existing services \cite{xliu2013}. It may be unfair to interpret a lower or higher reputation for an on-demand composition as the duration of the composition is not \textit{long enough} for accumulating proper feedback from users. 

\textit{Reputation bootstrapping} is an efficient process that performs an \textit{indirect assessment} of reputation when users' feedback is not available \cite{lieicsoc2017,DBLP:journals/internet/MalikB09}. Various \textit{reputation-related factors} such as the reputation of the service provider, community or interrelated services are used for the indirect assessment. The reputation of the service provider is used to predict the future performance of a new service \cite{DBLP:conf/wecwis/NguyenYZ12, lieicsoc2017}. Another approach \cite{DBLP:conf/wise/SkopikSD09} assumes that the reputation of a new service may follow the reputation trend of functionally similar services. In this paper, a service is considered \textit{similar} to another service if their \textit{functional properties} closely match. 

The \textit{on-demand} service composition is a natural phenomenon in popular service markets, e.g., Web, Cloud, Micro-services, and Business Process Management \cite{7401117}. The service environments are characterized by changing conditions in the explosive growth of new services, users' long-term QoS requirements, and QoS fluctuations in existing services \cite{Erl:2004:SAF:983556}. Usually atomic services collaboratively create on-demand value-added services following a composition topology. On-demand compositions are usually \textit{context-aware} and adapt to the changes in user requirements \cite{7401117,4279708}. If a single service does not have the functionality to meet a consumer's requirements, related services are composed on the fly \cite{xliu2013}. It is challenging to access the \textit{reputation of on-demand composite services} as they usually have little or no direct consumer feedback. 

\textit{Our objective is to build a reputation bootstrap framework for a new on-demand composite service.} We assume that the \textit{concrete implementations} of atomic services are already selected in a given on-demand composition topology. This assumption clarifies that we focus only on the reputation bootstrapping of on-demand compositions. \textit{The selection of the most reputable composition is outside the focus of the paper}. Note that the topology remains static, i.e., the topology does not change at the time of the reputation bootstrapping. The static topology could be used in real-world to assess the reputation of a new composition plan at the design level before its actual implementation \cite{6461455,xliu2013}. For example, a medical patient may require a highly reputable ``medical service" which is composed of ``doctor service" and a ``pathology service". Let us assume that an agent proposes a composition plan using ``Sydney doctor" and ``Doglus Pathology" services which are concrete implementations of ``doctor service" and a ``pathology service" respectively. An effective approach is required to determine the reputation of the composition plan on the fly which will help the patient to make an informed decision. 

\begin{figure}
   \centering
   \vspace{-6mm}
   \includegraphics[width=.4\textwidth]{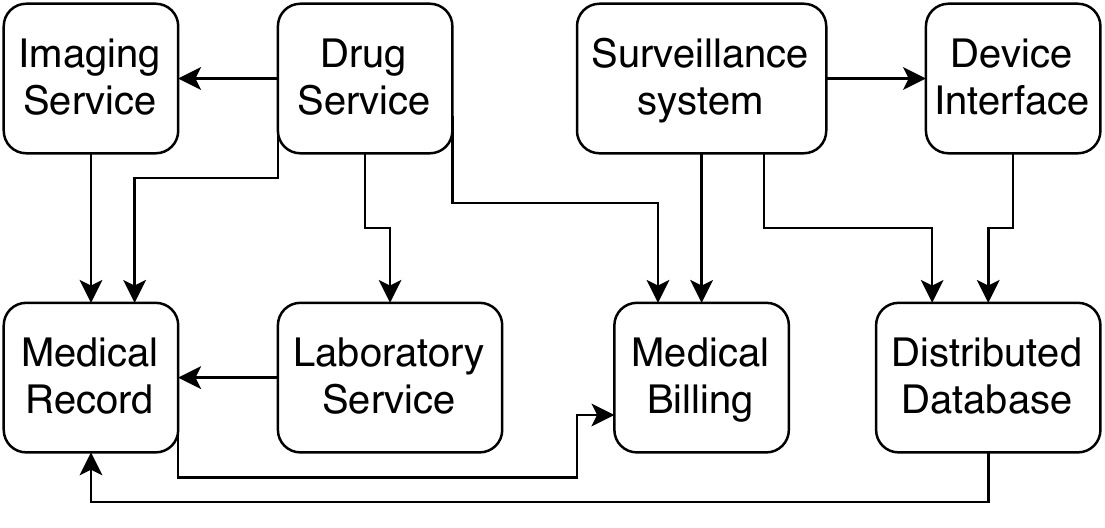} \vspace{-2mm}
   \caption{\small A composition topology for the Hospital Management System (HMS)}
   \vspace{-7mm}
   \label{fig:1}
 \end{figure}

Let us assume, a new software company plans to develop an open-source Hospital Management System (HMS) as a Github repository. A Hospital Management System (HMS) is enterprise software that manages and runs all the activities which are involved in running a healthcare facility. Fig. \ref{fig:1} represents the composition topology of an on-demand HMS. If two services are related in a composition, we use directional edges, i.e., arrows to represent such relationships. The arrow from a service $X$ to another service $Y$ represents that $X$ is a submodule of $Y$, i.e., $X$ should be executed first to complete the execution of $Y$. In Fig. \ref{fig:1}, the drug service (DS) requires electronic medical records (EMR), results from a laboratory (LS) and a surveillance system (SS) to prescribe medications. The surveillance system (SS) is dependent on the device interface (DI) and data sharing platform service (DP). Similarly, the medical billing service (MB) can generate invoices once the drug service (DS) and surveillance service (SS) finish their tasks for a specific patient. The software company identifies two possible implementations of HMS by composing the newly published open-source software available in Github, i.e., $HMS^{A}$ = \{GNU Health, Open-eobs, OpenEP, VTK, OpenELIS, OpenMBS, FHIRBase, OpenAPS\} and $HMS^{B}$ = \{FreeMD, echOpen, OpenHim, ITK, Bika, Billing Pro, Mirth Connect, Murgen\}. 
The software company wants to know which on-demand composition is \textit{more reputable}, i.e., $A$ or $B$?

The reputation bootstrapping of composite services differs from the reputation bootstrapping of single or atomic services. To the best of our knowledge, existing research is mostly focused on reputation bootstrapping of atomic services \cite{DBLP:conf/wise/SkopikSD09,DBLP:conf/atal/BurnettNS10,DBLP:conf/trustcom/JiaoLLL11}. In contrast, the reputation of a composite service depends on the aggregated reputation of its component services \cite{DBLP:journals/vldb/MalikB09}. \textit{Linear aggregation models} are proposed for bootstrapping composite services without considering the dynamic performance of component services in different contexts \cite{DBLP:journals/internet/MalikB09,DBLP:journals/vldb/MalikB09,lieicsoc2017}. However, aggregating reputation-related factors of component services is not straight-forward due to the \textit{service topology} and \textit{reputation interdependence} among component services. The reputation interdependence refers to the phenomenon that the reputation of a component service in a composition is dependent on its directional interactions with other component services. For example, GNU Health and Open-eobs in the composition $HMS^{A}$ may be individually more reputable than FreeMD and echOpen in the composition $HMS^{B}$. In such case, the composition $HMS^{A}$ is more reputable than the composition $HMS^{B}$ using the linear aggregation rules. However, we may find a different result when we consider the \textit{reputation interdependence} between reputation-related factors. There may exist an inverse reputation interdependence between open and custom standards. The community of GNU health usually works with custom standards and has expertise in desktop-based solutions. The community of Open-eobs usually work with open standards and has expertise in cloud-based solutions. Due to the community mismatch and inverse reputation interdependence, the composition $HMS^{A}$ may become less reputable than the composition $HMS^{B}$. To the best of our knowledge, the reputation interdependence and service topology are not considered to bootstrap composite reputation in the existing literature.

The reputation interdependence varies in different layers of the reputation-related factors. For example, a component service usually has multiple reputation-related factors (e.g., its provider, the community, and similar services) \cite{lieicsoc2017}. We use Conditional Preference Networks (CP-nets) \cite{boutilier2004cp} to model the \textit{relative importance} among reputation-related factors. Usually, there exists no prior knowledge about the relative importance of reputation-related factors in bootstrapping reputation of a service. Hence, we generate several candidate CP-nets to represent the reputation of a component service. \textit{As a result, the reputation bootstrapping problem is transformed into the composition of CP-nets}. There are two important steps in such compositions: 1) \textit{deriving composition rules}, and 2) \textit{selection of candidate CP-nets}. We determine the composition rules using the history of service invocations and the direction of reputation interdependence. For example, if a highly reputable service invokes a new service, the reputation of the new service may benefit and be treated at a high-level reputation. If a new service invokes a highly reputable service in a composition, it may not have a higher influence on the reputation of the new service.

We explore four different CP-nets composition approaches: a) brute-force approach, b) random approach, c) domain-best approach, and d) reduced topology approach. The naive or brute-force composition approach generates all possible combinations of candidate CP-nets and then applies linear composition rules, i.e., the mean reputation of all candidate services. This naive approach does not consider the \textit{composition topology} and the \textit{reputation interdependence} among the component services. The random approach selects CP-nets randomly from a set of candidate CP-net. The domain-best approach selects the CP-nets using domain knowledge, i.e., semantic annotations, semantic keywords  similarity, and the specification matching. We apply the reputation influence factor (RIF) of a component service as the heuristic to reduce the number of component services in a composition in the reduced topology approach. The reduced topology approach transforms the selection of optimal CP-nets as a sequential decision process. The proposed reduced topology approach applies reinforcement learning (Q-learning) using the reputation-interdependence as the optimal policies are not available supervised learning. As machine-learning approaches are proved to be efficient in solving a sequential decision process \cite{dorigo2016ant}, we focus on the model-free learning attempts to learn the optimal policy through trial-error (e.g., Q-learning \cite{watkins1992q}).

The key contributions of the paper are as follows:
\begin{enumerate}
\item A CP-nets based model to represent relative importance among reputation-related factors.
\item Determining reputation interdependence among the component services using the history of invocations.
\item A heuristic-based reduced topology approach for reputation bootstrapping using Q learning.
\end{enumerate}

\vspace{-4mm}
\section{Related Work}
The reputation bootstrapping of an atomic service is somewhat related to the generic cold-start problem present in recommendation systems \cite{Ricci2015}. In recommendation systems, the cold start problem refers to making recommendations for a new user or a new item that has no rating data. When the rating data is sparse, two users or items are unlikely to have common ratings. It is difficult to make recommendations using Collaborative Filtering (CF) in the cold start situation. Different approaches are proposed to solve the cold start problem, e.g., asking for explicit ratings, creating users' demographics and attribute-based techniques \cite{Kalloori:2017:ICS:3079628.3079696}.

Similar to the recommendation systems, the cold start situation may be observed to determine the reputation of a new service.  As there is no history of the service's performance, it requires reputation bootstrapping approaches. To the best of our knowledge, existing work on \textit{reputation bootstrapping focuses on atomic services}. The approaches are divided into three categories:

\begin{itemize}[leftmargin=*, nolistsep]
 \item \textit{Characteristic-based approaches:} This category focuses on  reputation-related factors \cite{lieicsoc2017} and uses \textit{inheritance} and \textit{referral} mechanisms to bootstrap reputation of a new service \cite{DBLP:conf/wecwis/NguyenYZ12}. A trust framework is proposed in \cite{XiongPeerTrust} that considers community context factors for establishing trust in the peer-to-peer electronic communities. The core idea in characteristic-based approaches is to find the reputation-related factors and mining association rules between the final reputation and their factors. However, these approaches do not explain how the reputation related factors are related to the composite reputation.  
 

\item  \textit{Guarantee-based approaches:} Service providers advertise a form of service level agreements (SLA) that are used to bootstrap the reputation \cite{DBLP:conf/atal/HuynhJS06}. A time-series prediction algorithm is proposed to determine the SLA of a new service in \cite{DBLP:conf/wecwis/NguyenYZ12}. A learning-based bootstrapping approach are proposed in \cite{DBLP:journals/eis/WuZL15} using Neural Network and Linear Regression respectively. 
Different frameworks are proposed to determine the reputation of cloud services where the SLAs are dynamic and change in the long-term period \cite{DBLP:journals/eis/WuZL15}. Setting up a service level agreement may not be possible for new service providers due to the lack of history or records on service performance. The SLA for a composite service may not be computed based on the aggregated SLAs of the corresponding component services. As different providers have different policies for their SLAs, it may not be possible to synchronize the policies in the runtime.

\item \textit{Trial-based approaches:} A service provider has trial periods to build its reputation according to the trail-base approaches \cite{DBLP:journals/internet/MalikB09}.  A machine-learning approach is proposed in \cite{DBLP:conf/icwe/YahyaouiZ11} to model the trust patterns of the service performance. The model predicts the services' future performance as the bootstrapped reputation. 
The single trust population statistics are associated with every new entity in some bootstrapping approaches \cite{DBLP:conf/icsoc/HuangLNFCT14}. However, these approaches are not feasible in composition scenarios as typically on-demand composite services do not have trial periods. 
\end{itemize}

As the qualitative reputation is transformed into a quantitative value to enable comparisons and rankings in a composition, graphical models with quantitative probabilities, e.g., Bayesian Network (BN), probabilistic CP-nets (PCP-nets) or GAI-networks \cite{cornelio2013updates} may be used for modeling reputation using the reputation related factors. Such models require historical data or observations to learn the right graphical representation and the corresponding probabilities to represent the reputation of a single component service. However, there exists no prior knowledge about the relative importance of reputation-related factors in bootstrapping reputation of a service. We also consider the reputation interdependence in the topology to design the reputation representation model. It is possible to represent reputation interdependence among component services by aggregating BNs, PCP-nets, or GAI-networks with the correlations among reputation related factors. However, the learned correlations may not be applicable to a new composition as it may have a completely new set of component services. The reputation interdependence is highly dynamic in new composition typologies. It is also challenging to learn correlations among reputation related factors for a new composition without historical data. Hence, BN, PCP-nets, or GAI-networks approaches are only effective for bootstrapping new composite services if a history pool of similar composition topology is used.

Reputation-related factors of each component services are determined using the layered-based approach of bootstrapping reputation for a single service \cite{lieicsoc2017,DBLP:journals/internet/MalikB09}. As required information, i.e., reputation-related factors are already generated, the cold-start situation does not apply to bootstrap the composite reputation. We focus on the existing approaches for \textit{QoS-aware service compositions} as reputation is a QoS or non-functional property of services. Linear Aggregation (LA) rules are usually applied on the QoS attributes (e.g., response time, cost, throughput) to form the global objective functions in the QoS-aware compositions \cite{ramirez2017evolutionary}. Such linear aggregations could not be applied to bootstrap the composite reputation for two main reasons: 1) initial reputations of the component services are not available (they require \textit{bootstrapping} approaches) and 2) composite reputation is \textit{non-linearly correlated} with the composition topology and reputation-related factors \cite{lieicsoc2017}. Regression analysis is usually applied to predict future performance using historical data. A Linear Regression (LR) approach is proposed to predict the popularity of a Github repository \cite{Borges:2016:PPG:2972958.2972966}. Such approaches may not be applicable for on-demand compositions as they do not consider the correlations among the composite reputation, composition topology and the influence of reputation-related factors.

\vspace{-4mm} 
\section{Reputation Bootstrapping Framework}
\begin{figure*}
	\vspace{-7mm}
   \centering
   \includegraphics[width=.9\textwidth]{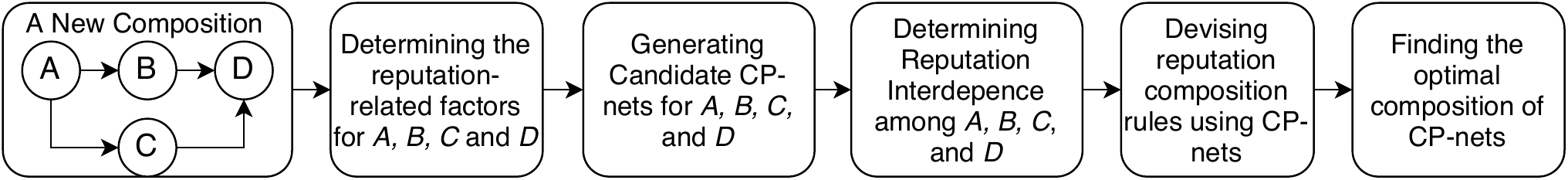} 
   \caption{The framework to bootstrap the reputation of a composite service }
   \vspace{-5mm}
   \label{fig:framework}
 \end{figure*}


Our target is to design a reputation bootstrap framework using the composition topology and the reputation interdependence among the reputation-related factors. The key reputation-related factors \cite{lieicsoc2017} are identified as follows:
\begin{itemize}[leftmargin=*, nolistsep]
\item \textit{Provider:} A reputable provider has a higher probability to offer the expected service \cite{lieicsoc2017}. In the context of Github, a contributor to a repository may be regarded as a provider. 

\item \textit{Community:} A service belonging to a reputable community has a higher probability of providing good services \cite{XiongPeerTrust}. In Github, a community is the organizer of a repository. 

\item \textit{Interrelated Services:} Services are often consumed in a bundle. In the context of medical services, a pathology service may be recommended by the doctor. Hence, the reputation of the pathology service may be propagated to the reputation of the doctor. In the context of Github, two projects $A$ and $B$ may be used by the same community of users (They are called watchers in the context of Github). The project watchers usually rate similar projects. If there are two interrelated projects $A$ and $B$, their interrelation can be computed using the Mutual Information score (MI-Score) using the number of their share watchers ($X$ and $Y$ refer to the Number of \textit{watchers} in $A$ and $B$ respectively, $Z$ refers to the number of shared \textit{watchers} in $A$ and $B$) and $I$ refers to number of invocations among $A$ and $B$) 
in Equation \ref{eq:s1}. If they are highly interrelated, the reputation of $A$ may be used to bootstrap the reputation of $B$ or vice versa. Note that the efficiency of a metric depends on the domain of the service and available information. For example, Hybrid Similarity (HSim) is used to link similar web pages or scientific papers based on keywords \cite{7792740}. 
\vspace{-1mm}
\begin{equation}
\label{eq:s1}
\text{Interrelation} (A,B) = log_{2}\frac{Z \times I}{X \times Y}
\end{equation}

\item \textit{Service meta-information:} The reputation of the service can be associated with the meta-information. In the context of Github, a repository can contain meta-information such as documentations, Wikis, Issue tracking and Pull requests. The quality of such meta-information may be used to assess the reputation of a newly published repository. 
\end{itemize}


The reputation-related factors may have different degrees of importance in representing a service's reputation. For example, the reputation of a provider may be more important than the other reputation-related factors. In medical services, the educational backgrounds and experience of a doctor may be more important than the hospital location to assess the doctor's reputation. In some cases, the community may be the most important reputation-related factor. In business services, the reputation of the organization is usually more important than their employees' educational backgrounds. \textit{We assume that there exists no historical dataset containing quantitative values of reputation-related factors.} As an on-demand composition consists newly published component services, it is unlikely that a historical dataset could be formed for the new services in a short-period of time. We identify the following challenges:

\begin{itemize}[leftmargin=*, nolistsep]
    \item \textit{Qualitative representation of reputation of a new service:} Due to lack of direct feedback of user ratings, bootstrapped reputation is usually represented using reputation-related factors. As reputation is qualitative, it is natural to apply qualitative graphical models to capture the relative importance of reputation-related factors to represent reputation. To the best of our knowledge, existing bootstrapping approaches do not explore the application of graphical models to represent reputation.
    
     \item \textit{Reputation Composition Rules using the Reputation Interdependence:} The Linear Aggregation (LA) \cite{ramirez2017evolutionary} may not applicable in a composition as it considers each component services independently. However, the component services are dependent and there exists reputation interdependence among their factors. The key challenge is to quantitatively determine reputation interdependence and to apply weighted influence factor to devise reputation composition rules using graphical models.
     
\end{itemize}

In contrast to the data-driven approach, flexible graphical models are proposed to represent the qualitative preferences of the users in a composition environment \cite{sajibicsoc2016}. CP-nets \cite{boutilier2004cp} are represented using the dependency graph that is associated with the conditional preference table to hold the relative ordering of qualitative preferences in a composition. To the best of our knowledge, existing approaches do not describe how CP-nets could be used to represent the reputation of a service. As we do not consider a history pool of composition topology and their composite reputation values for the supervised-learning, we design the bootstrapping framework using the flexible graphical model, i.e., CP-nets. In future, we will explore model-based approaches to determine the composite reputation when a history of composition topologies is available.

Different CP-nets could be constructed varying the relative importance due to no prior knowledge about the relative importance of reputation-related factors. Each component service may have multiple of candidate CP-nets. The challenge is the selection of the optimal set of CP-nets for each component service. As the solution space is relatively large, our objective is to find heuristics to reduce the search space in run-time with higher accuracy. Fig. \ref{fig:framework} depicts the high-level architecture of the framework. We identify the following key steps:
\begin{itemize}[leftmargin=*, nolistsep]
\item Step 1 - \textit{Input}: the framework will take a composition of component services as input. We consider the following types of composition topologies: a) \textit{sequential}, b) \textit{parallel}, c) \textit{loop}, and d) \textit{hybrid} - combinations of sequential, parallel and loop \cite{7060671,7027801}. For example, a parallel composition of four component services, $A$, $B$, $C$, and $D$ are considered as the input in Fig. \ref{fig:framework}.   
\item Step 2 - \textit{Determining Reputation-related factors for component services:} we assume that \textit{the reputation-related factors are mined by existing approaches or identified by an expert for an abstract service}. For example, the \textit{provider} and the \textit{community} may be the identified reputation related factors for the component service $A$. The \textit \textit{interrelated services} and the \textit{meta-information of service popularity} may be the identified reputation related factors for the component services $B$ and $C$ respectively. We incorporate data-driven approaches \cite{lieicsoc2017} or domain experts to identify the reputation-related factors and only focus on applying CP-nets to bootstrap the composite reputation (i.e., steps 3 to 5) in this paper. Note that domain expert is not a mandatory requirement to solve step 2. 
\item Step 3 - \textit{Generating candidate CP-nets that represent reputation}: One critical issue using CP-nets to represent reputation is that it does not quantify the intensity of the reputation-related factors. One solution is to apply weights on the edges of CP-nets \cite{wang2012wcp} or probabilities in PCP-nets \cite{cornelio2013updates}. \textit{As we do not consider historical data on relative importance, determining the weights or probabilities with historical data is out of this research scope.} We generate candidate CP-nets with random preferences among reputation related factors. For example,  \{$CP_{a1}, CP_{a2},....,CP_{aN}$\} is the set of candidate CP-nets for $A$. Similar candidate services are also specified for $B$, $C$, and $D$.
\item Step 4 - \textit{Determining Reputation Interdependence among the component services}: We compute the reputation dependence using the history of invocation of services. For example, if $B$ is mostly invoked by $A$ in different compositions over a period of time, it may be deduced that $A$ is highly dependent on $B$.

\item Step 5 - \textit{Devising CP-nets composition rule to compute the reputation}: We focus on computing the composite reputation of a composition of CP-nets using the reputation interdependence. For example, the composition,  \{$CP_{a1}, CP_{b2},CP_{c3}, CP_{d4}$\} will produce a composite reputation for the input composition in Figure \ref{fig:framework}.     
\item Step 6 - \textit{Transforming composite reputation bootstrapping problem into a CP-nets composition problem}: we select the optimal candidate CP-nets and computing the final reputation value. In Figure \ref{fig:framework}, our target is to learning the selection of the optimal set \{$CP_{ai}, CP_{bj},....,CP_{ck}$\} that optimizes the objective function, i.e., reducing the error between computed composite reputation using composition rules and the actual composite reputation (ground truth). 
\end{itemize}
The detailed implementation of steps 3 to 6 are structured in the paper as follows. We define CP-nets to represent reputation (step 3) in Section 4. The step 4 in Fig. 2, i.e., determining the reputation interdependence among component services is discussed in section 5.1. Section 5.2 describes the step 5, i.e., devising the reputation composition rules using CP-nets and the reputation interdependence. Section 5.3 describes how the composition rules (step 5) are applied as the objective function in the step 6 to find the optimal composition of reputation-related CP-nets.

\vspace{-2mm}
\section{Reputation Modeling using CP-nets}
Although \textit{provider}, \textit{community}, \textit{interrelated services} and \textit{service meta-information} are most common reputation-related factors, we generalize that there may be a set ($RF$) of $N$ reputation-related factors ($F_{l}$) for $S^{A}_{I}$. We represent it as $RF(S^{A}_{I}) = \{F_{l} \;| \;\forall \;l \;\epsilon \;[1,N]\}$. First, our target is to bootstrap the reputation of a concrete or component service.  $S^{C}_{IJ}$ is the $J^{th}$ component service of the abstract service, $S^{A}_{I}$. We bootstrap the reputation from $RF$. 

\begin{figure}[t!]
   \centering
   \vspace{-7mm}
   \includegraphics[width=.48\textwidth]{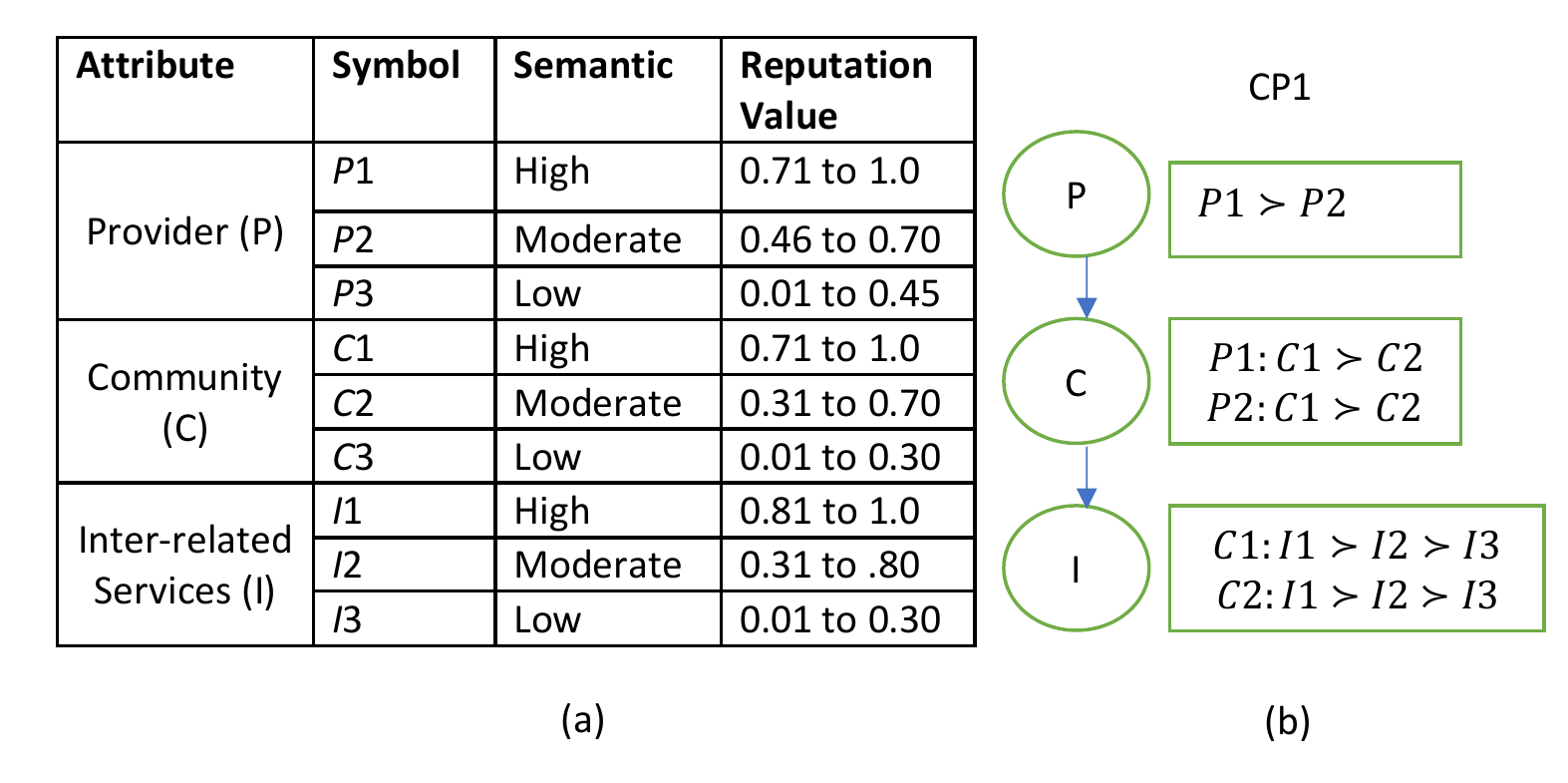}
   \vspace{-3mm}
   \caption{\small (a) Three semantic segmentation of reputation values for reputation-related factors, and (b) A candidate CP-net to represent the reputation of a component service }
   \vspace{-4mm}
   \label{fig:2}
 \end{figure}

We define CP-nets for representing reputation ($RCP$) as 4-tuple $<RF, ST, G, CPT>$:
\begin{itemize}[leftmargin=*, nolistsep]
\item  $RF$ represents a set of reputation-related factors. Typical factors are Provider ($P$), Community ($C$), Interrelated Services ($I$), and Meta-information ($M$).
\item $ST$ represents the semantic interpretations over ranges of a reputation factor $F_{i}$. In Fig. \ref{fig:2}(a), 0.8-1.0 reputation is interpreted as a ``high'' reputation. \textit{In this paper, we do not focus on finding an optimal semantic segmentation, rather we focus on how a predetermined semantic interpretation could be efficiently leveraged in bootstrapping reputation. }     
\item $G$ is a directed graph where each node is a reputation-related factor in $RF$. The arc from a node ($N_{1}$) to another node $N_{2}$) represents that the \textit{reputation influence} of $N_{1}$ is greater than $N_{2}$. If the Provider ($P$) is more important than the Community ($C$) in bootstrapping reputation, there will be an arc from $P$ to $C$ (see Fig. \ref{fig:2}(b)). 
\item The nodes in $G$ are associated with conditional preference tables $CPT(F_{i})$ for each $F_{i} \in RF$. Each conditional preference table $CPT(F_{i})$ depicts the conditional reputation influence of the values in $F_{i}$ where the parent nodes are given. For example, if there is an arc from Provider ($P$) to Community ($C$), the semantic interpretations of $C$ is dependent on the semantic interpretations of $P$. The CPT $(P1:C1 \succ C2)$ in Fig. \ref{fig:2}(b) implies that if the reputation of the provider ($P$) is ``high", the ``high" reputation in the community is highly influential than the ``moderate" reputation of a community.
\end{itemize} 

A CP-nets which is acyclic in nature can generate the total ordered ($\succeq$) ranking of the preferences or configurations \cite{boutilier2004cp}. Hence, we generate the relative order of the set of reputation-related factor configurations in  $RCP$ using induced graph approach in \cite{Wang2016124}. Let us assume $CRF_{i}$ is the $i^{th}$ reputation configuration. $<P1, C1, I1>$ is such configuration in Fig. \ref{fig:2}(b). According to \cite{boutilier2004cp}, $CRF_{i} \succeq CRF_{j}$ means that a configuration $CRF_{i}$ is equally or more important than $CRF_{j}$ to bootstrap reputations. We use $CRF_{i} \succ CRF_{j}$ when the reputation influence of $CRF_{i}$ is higher than $CRF_{j}$.  Relative influence between two configurations could be indifferent, i.e., $CRF_{i} \sim CRF_{j}$.

In Fig. \ref{fig:2}(b), we present a CP-nets to represent the reputation of a service. In $CP1$, ``provider'' of a service is the most important reputation-related factor, followed by ``community'' and ``interrelated services''. In $CP1$, the ``high'' reputation of a provider has a higher importance than the ``moderate'' reputable provider, i.e., $P1 \succ P2$. Note that, the ``low'' reputable provider ($P3$) is not considered in $CP1$. It means that if the reputation of a provider is very low, the reputation of the corresponding service is automatically considered as low. Once the reputation of the provider is determined, the next thing is to explore the importance of the community using the provider's reputation. In the CPTs of $CP1$, ($P1:  C1 \succ C2$) means that the higher reputation of the community may bootstrap higher reputation of the service than the ``moderate" reputable community if the corresponding provider's reputation is also higher. Finally, the reputation of the interrelated services is considered based on the reputation of the provider and the community. In the CPTs of $CP1$, ($C1:  I1 \succ I2 \succ I3$) means that a higher reputation of the interrelated services may bootstrap higher reputation of the service than the moderate and lower reputable interrelated services if the corresponding provider's and communities reputation are also higher. Following the relationships in $CP1$, we can state that the higher reputation of the service is represented by the configuration by the configuration ($P1,C1,I1$) and the lower reputation of the service is represented by the configuration by the configuration ($P2,C2,I2$). It may not be possible to have prior knowledge about the relative importance of reputation-related factors. It means that the reputation of the service can be represented in different candidate CP-nets. For example, a CP-net could be formed where ``community'' of a service is the most important reputation-related factor, followed by ``provider'' and ``interrelated services''.  

The configurations or reputation influences in $RCP$ are partially ordered based on their importance. We apply the \textit{induced preference graph} \cite{boutilier2004cp} to represent these orders. The graph is an acyclic directed graph where each node is the complete configuration or the reputation influence which is given in each the CPTs. The graph is constructed from a $RCP$ by the pairwise comparisons of the configurations \cite{Wang2016124}. Fig. \ref{fig:f3}(a) depicts the induced preference graph of $CP1$. As the $CPT(CP1)$ states $P2: C1 \succ C2$, and $C2: I1 \succ I2 \succ I3$, There is an edge from $(P2, C2, I3)$ to $(P2, C2, I2)$ and the next edge starts from $(P2, C2, I2)$ to $(P2, C2, I1)$ considering the \textit{ceteris paribus} preference statements. 
Each node in the induced preference graph is associated with a score. A higher score is mapped with a higher reputation value for the service. The root node is given the score value 1. Starting from the root node, the graph is traversed in a Breadth-First Search (BFS) manner. Fig. \ref{fig:f3}(a) depicts a total ordering of induced graph. The score of $(P2, C2, I3)$ is 1 and the score of $(P1, C1, I1)$ is 11 in Fig. \ref{fig:f3}(b). Fig. \ref{fig:f3}(b) depicts a partial ordering of induced graph. 
Two mapping methods are used to transform the qualitative preferences into a quantitative reputation value:

\begin{figure}[t!]
   \centering
   \vspace{-7mm}
   \includegraphics[width=.55\textwidth]{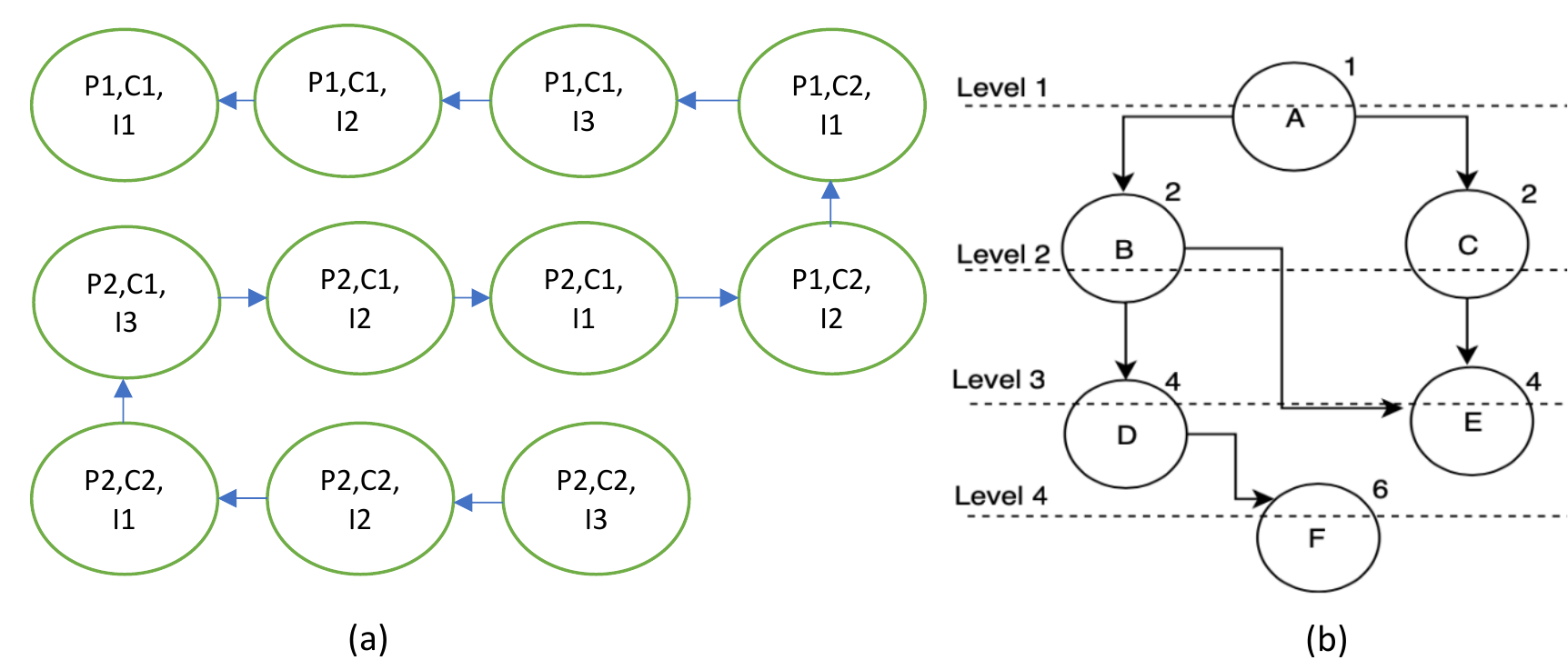}
   \vspace{-2mm}
   \caption{(a) A total ordered induced ranking of CP1, (b) A partial ordered induced ranking}
   \label{fig:f3}
   \vspace{-5mm}
 \end{figure} 

\begin{itemize}[leftmargin=*, nolistsep]
\item \textit{Mapping with Uniform distribution:} We assume that the $N$ score values are uniformly distributed in the range [0,1]. The reputation value is proportional to its score value. The mapped reputation range ($RR$) for the $i^{th}$ score is calculated as follows:
\vspace{-2mm}
\begin{equation}
\small
RR(i) = [(i-1) \times \frac{1}{N}, i \times \frac{1}{N}]
\end{equation}  

\item \textit{Mapping with Normal distribution:} The importance of the factors can be delineated at multiple levels or groups as described in \cite{Wang2016124}. The relative importance (weight) decreases when the level increases which reflects the extent to which an attribute is more important than another. For example, an attribute $F_{1}$ can have a very high importance value to another attribute $F_{2}$ at level 1 or group 1, $F_{2}$ can have a moderate importance value to another attribute $F_{3}$ at level 2 or group 2, and $F_{3}$ can have low importance value to the attribute $F_{2}$ at level 3 or group 3. It means that higher and lower reputation group should have a lower number of mapped score values than the moderate reputation values. Such distribution is closely matched with normal distribution. As the weights are not explicitly specified, the $N$ score values are normally distributed to different groups in the range [0,1]. The score values are distributes in 5 groups: 1st: ($0, .021 \times N$), 2nd: ($.021 \times N, .156 \times N$), 3rd: ($.156 \times N, .836 \times N$), 4th: ($.836 \times N, .97 \times N$), 5th: ($.97 \times N, N$).  If the mean value of the reputation is $\mu$ and standard deviation $\sigma$, the mapped reputation range ($RR$) for the $i^{th}$ score is calculated in Equation \ref{eq:repucpcal}. Here, $l$ is the group number of the $i^{th}$ score, $n$ is the total number of scores in group $I$, and $o$ is the ranking of the $i^{th}$ score in group $I$.
\begin{equation}
\label{eq:repucpcal}
\small
RR(i) = [(o-1) \times \frac{1}{n} \times I \times \sigma, o \times \frac{1}{n} \times I \times \sigma]
\end{equation}  
\end{itemize}
\vspace{-5mm}
\section{Composite Reputation Bootstrapping}
\subsection{Reputation Interdependence in a Composition}
We consider the history of interactions of component services in a composition. Note that, it is not the correlations among reputation-related factors. Let us assume, two component services $S^{C}_{ij}$ and $S^{C}_{mn}$ has been invoked $N$ and $M$ times in the interval $T_{t}$. Both $S^{C}_{ij}$ and $S^{C}_{mn}$ are invoked together $P$ times in the same interval. If $N \sim P$, it states that most of the time when $S^{C}_{ij}$ are invoked, $S^{C}_{mn}$ are most probably invoked. However, the vice-versa is not true if $M << P$. We can state that the reputation of $S^{C}_{ij}$ is highly dependent on $S^{C}_{mn}$, but $S^{C}_{mn}$ is not highly dependent on $S^{C}_{ij}$. The reputation dependence ($D$) of a component service, $S^{C}_{mn}$ on $S^{C}_{ij}$ in an interval $t_{0}$ is calculated as follows:
\vspace{-2mm}
\begin{equation}
\small
D_{S^{C}_{mn}, S^{C}_{ij}}(t_{0}) = \frac{\text{Number of invocations of } S^{C}_{mn}}{\text{Combined invocations of }S^{C}_{mn} \text{, }S^{C}_{ij}} 
\end{equation} 

\begin{figure}
   \centering
   \vspace{-8mm}
   \includegraphics[width=.3\textwidth]{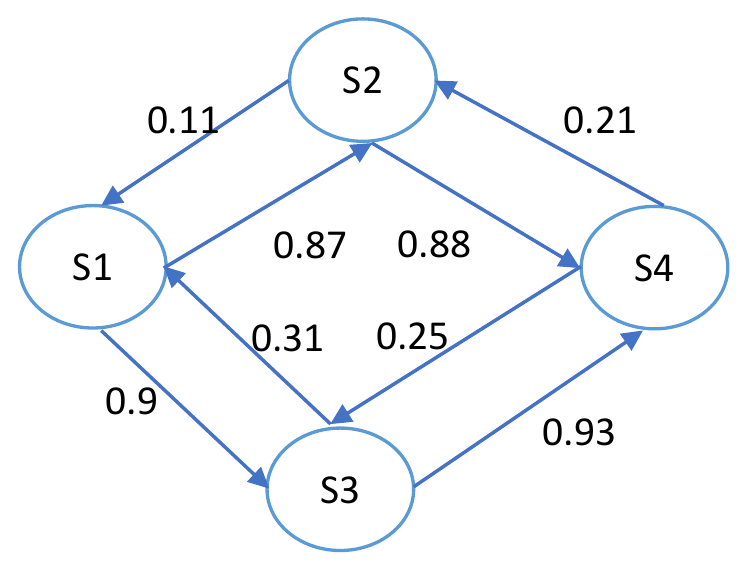}
    \vspace{-2mm}
   \caption{Reputation interdependence in a topology}
   \vspace{-2mm}
   \label{fig:4}
   \vspace{-3mm}
 \end{figure}
 
It may not be possible to find the history of invocations statistics In the real-world. As we consider Github as the composition environment, we treat the ``watchers'' as evidence of invocations. Hence, if a significant number of watchers of a repository also watch another repository, the reputations are intertwined. The reputation dependence between the two services is also temporal. 
The pairwise historical dependence time-series of all component services are represented in the following matrix ($MD$):

\small \[MD(t)=\begin{bmatrix}
D_{11}(t)&D_{12}(t)&\cdots &D_{1n}(t) \\
D_{21}(t)&D_{22}(t)&\cdots &D_{2n}(t) \\
\vdots & \vdots & \ddots & \vdots\\
D_{(n-1)1}(t)&D_{(n-1)2}(t)&\cdots &D_{(n-1)n}(t)\\
D_{n1}(t)&D_{n2}(t)&\cdots &D_{nn}(t)
\end{bmatrix}\]
\normalsize
$MD$ could not be directly used to bootstrap the reputation. If the composition is performed at the interval $T_{t+1}$, we need to predict the relationship dependence of the component services. As the historical dependence is represented as time series ($D(t)$), we use one-step time-series prediction technique \cite{zhang2003time} to calculate $MD(T_{t+1})$. 
Fig. \ref{fig:4} depicts a reputation dependency in a topology. Here, the arc with the weighted value 0.87 from $S1$ to $S2$ means that the reputation of $S1$ is highly dependent on the reputation of $S2$. However, the arc with the weighted value 0.11 from $S2$ to $S1$ means that the reputation of $S2$ is loosely dependent on the reputation of $S2$. 
\vspace{-4mm}
\subsection{Reputation Composition Rules} 
\label{sec:comrep}
The reputation dependence matrix $MD(T_{t+1})$ generates important knowledge about how the reputation of individual component services influence the reputation of the composition. We identify four types of influence: 

\begin{itemize}[leftmargin=*, nolistsep]
\item \textit{Highly Influential:} In a topology, this type of service has higher influence weights on most of the incoming edges but has lower influence weights on most of the outgoing edges. 
In Fig. \ref{fig:4}, $S4$ is a high influential service.

\item \textit{Moderately Influential:} This type of service has a higher influence weight on some of the incoming edges and has higher influence weights on some of the outgoing edges. 
In Fig. \ref{fig:4}, $S2$ and $S3$ are moderate influential services. 

\item \textit{Lowly Influential:} In a topology, this type of service has higher influence weights on most of the outgoing edges but has lower influence weights on most of the incoming edges. 
In Fig. \ref{fig:4}, $S1$ is a low influential service.

\item \textit{Neutral:} In a topology, this type of service has similar influence weights on most of the incoming edges and outgoing edges. 
The reputation of this service should have a higher influence in a composition.
\end{itemize} 

Based on our observations, we calculate the reputation influence factor of a component service ($X$) in Equation \ref{eq:rif}. The Average outgoing weight, $AOW$ indicates the total weighted dependency of other component services to the $X$ in the composition topology.  The maximum incoming weight, $IW$ indicates the maximum weighted dependency of $X$ to other component services in the composition topology. The reputation influence factor, $RIF$ indicates the ratio between $AOW$ and $IW$ in the range [0,1].
\vspace{-2mm}
\begin{align}
\small
\label{eq:rif}
&AOW = \frac{x_{i1}+x_{i2}+....+x_{in}}{n}\\\notag
&IW = max(x_{1i},x_{2i},.....,x_{ni})\\\notag
&\text{If } AOW < IW, \text{influence factor, } RIF(X) = \frac{AOW}{IW}\\\notag
&\text{If } AOW \geq IW, \text{influence factor, } RIF(X) = \frac{IW}{AOW} 
\end{align}   

Let us assume that the CP-nets, $CP_{ij}$ is already selected to represent the reputation of component service, $S^{C}_{ij}$. $RR(CP_{ij}, S^{C}_{ij})$ is the mean reputation value using equation \ref{eq:repucpcal}. Hence, the reputation of the composite service ($CR$) is calculated using the aggregated reputation of each component service with their weighted influence factor as follows:

\begin{equation}
\small
\label{eq:Rcal}
CR = \frac{\sum RIF(S^{C}_{ij}) \times RR(CP_{ij}, S^{C}_{ij})}{\sum RR(CP_{ij}, S^{C}_{ij})} 
\vspace{-2mm}
\end{equation}


\vspace{-2mm}
\subsection{CP-nets Composition Approaches}
\vspace{-1mm}
\subsubsection{Random Approach}
We select a CP-nets randomly from the set of candidate CP-nets for a component service. The reputation related-factors are predetermined for the component service. The reputation of the component service is determined by the Equation \ref{eq:repucpcal}. This approach is performed for all the component services. The composition topology provides the reputation influence factors of all component services. Once, all their reputation is calculated, the composite reputation is calculated using the Equation \ref{eq:Rcal}. 
\vspace{-3mm}
\subsubsection{Domain-best Approach} 

The domain or community reputation is the aggregated reputation of similar component services \cite{DBLP:journals/internet/MalikB09}. There exist several approaches to find similar services using the semantic annotations, semantic keywords similarity, and the specification matching \cite{Jian1531265}. \textit{Finding the optimal similar services in a domain is out of the focus of this work}. We apply the semantic keywords similarity approach \cite{Jian1531265} to find similar services. Note that other approaches can be also applied to find the community. The following procedure selects the best CP-nets for the given component service:

\begin{enumerate}[leftmargin=*, nolistsep]
\item We do not consider the similar component services that are not directly rated by the users. $N$ similar component services are found in the domain which is directly rated.
\item Select a CP-nets and calculate the reputations of the component services using the Equation \ref{eq:repucpcal}. Calculate the average difference between the calculated reputation and the ground reputation.
\item Repeat the previous step for all candidate CP-nets for the given component service. Finally, return the CP-nets which generates the lowest difference between the calculated reputation and the ground reputation as the best CP-nets for the component service. 
\end{enumerate}
 
We select the best CP-nets for all the component services in a composition. Then, the Equation \ref{eq:Rcal} is applied to calculate the composite reputation. 
\vspace{-4mm}
\subsubsection{Brute Force Approach}
\label{sec:brute}

The brute force approach creates all the combinations of candidate CP-nets. We calculate the probability of $R_{i}$ to be the optimal bootstrapped reputation. First we segment the reputation range into $\delta$ segments.  The range of the $i^{th}$ segment ($i\leq \delta$) is $[(i-1) \times \frac{1}{\delta}, i \times \frac{1}{\delta}]$. $Freq (i)$ denotes the frequency of reputation values on the $i^{th}$ segment. We calculate the $Freq (i)$ as follows:

\begin{enumerate}
\item Set $Freq (i) = 0$
\item For each $R_{i}$ in $0$, if $R_{i}$ in $[(i-1) \times \frac{1}{\delta}, i \times \frac{1}{\delta}]$, increment $Freq (i)$ by one.
\end{enumerate}

The probability of the optimal reputation is mapped into the segment $i$ is $\frac{Freq (i)}{M^N}$. We can select the segment $i$ only if $\frac{Freq (i)}{M^N}$ is relatively larger than $\frac{Freq (j)}{M^N}$ for all other segments $j$. If it is not, we increase the value of $\delta$ and start the procedure again.

\vspace{-2mm}
\subsubsection{Reduced Topology Approach}

\begin{figure*}
   \centering
   \vspace{-9mm}
   \includegraphics[width=.65\textwidth,height=.22\textwidth]{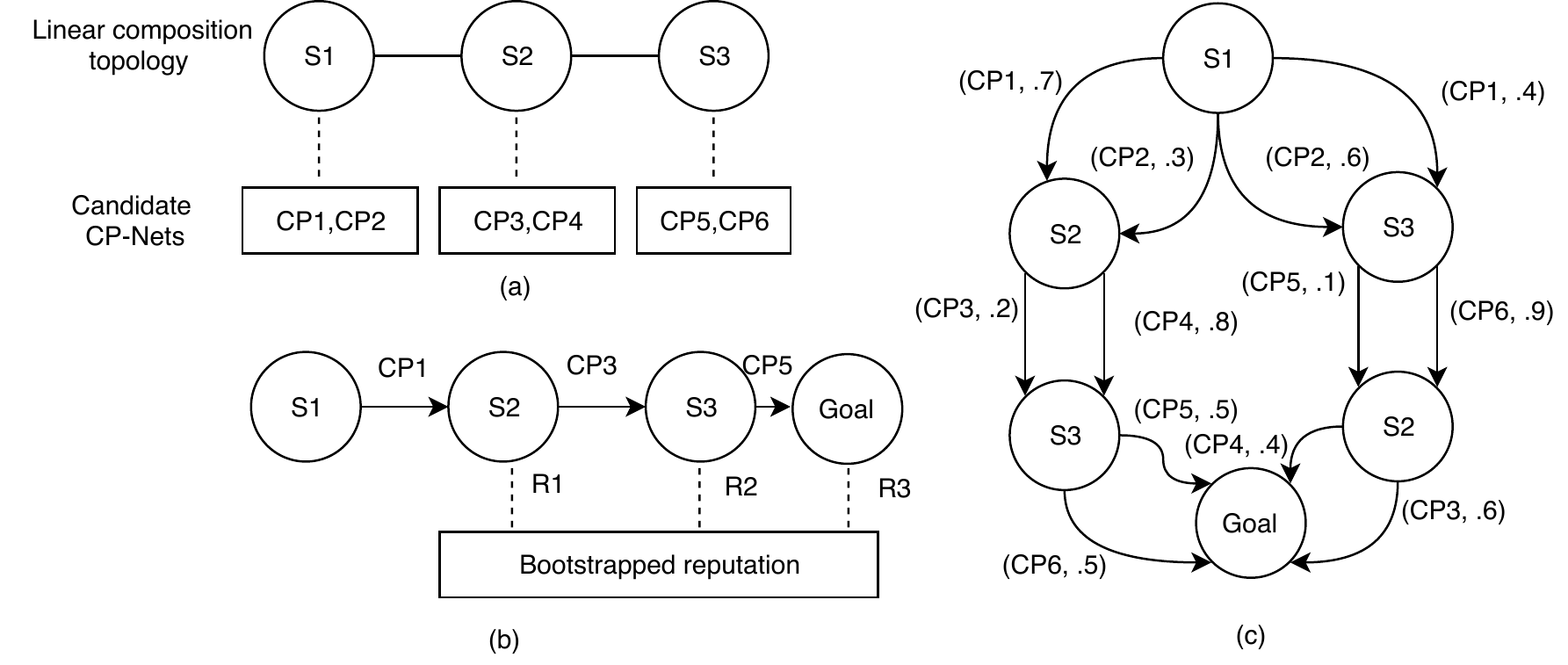}
   \vspace{-2mm}
   \caption{\small (a) A linear composition topology with candidate CP-nets, (b) A linear decision process, (c) A MDP with transition probabilities in the selection process \normalsize}
   \label{fig:f6}
   \vspace{-5mm}
 \end{figure*} 
 
It is not computationally efficient to create $M^{N}$ combinations of CP-nets if both $M$ and $N$ are a relatively large value. The workaround is reducing the number of component services $M$ to a significantly lower value of $m$. We apply the reputation influence factor ($RIF$) of a component service as the heuristic to reduce the number of component services $M$.  A low $RIF$ value means that the service is more reputation-dependent on other services rather than other services that are reputation-dependent on it. Hence, such component services could be removed from the set of component services as they should have a lower Influence on the overall reputation of a composition. $TIF$ is the threshold value to be included in the topology.  If $RIF(S^{C}_{ij}) < TIF$, the corresponding abstract service $S^{A}_{i}$ are removed from the topology. Once the topology gets reduced, we have the following options:
\begin{enumerate} [leftmargin=*, nolistsep]
\item Applying the brute force approach with the reduced topology (described in Section \ref{sec:brute}).
\item Applying a ML approach with the reduced topology.  
\end{enumerate}

We transform the reputation composition problem, i.e., the selection of CP-nets, as a \textit{sequential decision process}. We start with any component service in the topology and select a candidate CP-net to represent its reputation. The selection of the candidate CP-nets for the next component service depends on the previous selections of CP-nets as there may exist reputation interdependence. 

A composition topology is entitled as ``\textit{new}'' when it does not match with history. For example, a topology with three component services is new if each topology in history has less than or greater than three component services. Even different topologies with a similar number of component services can have different reputation-related factors. It is quite challenging to apply a model-based approach to bootstrap the reputation of new composite services if they are not \textit{exactly fit} to the model. For example, let us assume that a BN is modeled to predict the reputation of a composition topology with $N$ number of component services and $M$ number of reputation-related factors. The BN may not be applicable to predict the reputation of a new composition topology with a different X number of component services and Y number of reputation-related factors. As the composition environment is dynamic, model-free reinforcement learning, e.g., Q-learning is usually more applicable than the model-based learning algorithms \cite{sutton1998reinforcement}. The Q-learning treats each new composition topology as a new environment and learns the optimal selection of CP-nets through multiple interactions with the environment.

A sequential decision process could be modeled in different approaches such as Multi-Armed Bandit (MAB), Markov Decision Process (MDP), or Partially Observable Markov Decision Process (POMDP) \cite{chen2013combinatorial}. We model the optimal selection of CP-nets as a Markov Decision Process (MDP). We define the selection of a CP-net as an action. Each action generates a current reward, i.e., the fractional reputation. The goal of the MDP is to find the best sequence of actions that maximizes the total reward, i.e. the optimal bootstrapped reputation value \cite{chen2013combinatorial}. Note that, MAB or POMDP could also be used in our context as they are special cases of MDP. For example, MAB is a special case of MDP that has only one state \cite{chen2013combinatorial}. However, we select MDP as the general sequential decision process for bootstrapping the composite reputation. We will compare the performance of MAB and MDP in future work. We will explore the efficiency of MAB and POMDP over MDP in future work. A simple linear topology is depicted with three abstract services $\{S1,S2, S3\}$ in Fig. \ref{fig:f6}(a). Each service has candidate CP-nets: $\{S1:(CP1,CP2), S2:(CP3,CP4), S3:(CP5,CP6)\}$.  Fig. \ref{fig:f6}(b) depicts how the selection of CP-nets could be transformed as a decision process. The selection of $CP1$ for $S1$ leads to the selection of $CP3$ for $S2$ and so on. The final selection is performed in $S3$ with $CP5$ when decisions for all the nodes are made. Each selection of CP-nets computes a ``fractional reputation" (e.g., $R1,R2,R3$) which can be treated as the \textit{current reward} and the final bootstrapped reputation is calculated using Equation \ref{eq:Rcal} (Fig. \ref{fig:f6}(b)) with the aggregated current rewards.   

We model the MPD for the CP-nets composition for bootstrapping reputation as a 5-tuple $Com\_R=<S, A_{s}, s_{0}, P(. , .),R(. , .)>$. Here, $S = \{S^{A}_{1},S^{A}_{2},....,S^{A}_{m}\}$ is a finite set of abstract services in the reduced topology. We treat the corresponding component services in a topology as states in the MDP. $A_{S^{A}_{s}}$ is the finite set of actions, i.e., the candidate CP-nets which are available for the component services of $S^{A}_{s}$. $s_{0}$ is the starting state where the first selection of a CP-nets is performed for the corresponding component service. The selection of next CP-nets for reputation bootstrapping depends on these previous selections. $P_{CP_{a}}(S^{A}_{s}, \acute{S^{A}_{s}})$ is the probability to reach the optimal result by choosing a particular CP-nets $CP_{a}$ in the transition from $S^{A}_{s}$ to $\acute{S^{A}_{s}}$.$R_{a}(S^{A}_{s}, \acute{I_{s}})$ is the immediate bootstrapped reputation that the CP-nets $CP_{a}$ for $S^{A}_{s}$ produces after transitioning to the next state $\acute{I_{s}}$. We can calculate the composite reputation using Equation \ref{eq:Rcal}.

The transition to the next abstract service does not necessary to the order of the topology. The start state can be chosen randomly. We visualize a transition graph in an MDP for the linear topology in Fig. \ref{fig:f6}(c). Let us assume that the starting state is $S1$. The transition could happen either to state $S2$ or state $S3$ after selecting a CP-nets for $S1$. The transition is performed to $S2$ after selecting $CP1$. As the selection for $S1$ is already finalized, any future transition could not be performed to $S1$. The only transition possible from $S2$ is to $S3$. The action in $S2$ is $CP3$. Finally, the action in $S3$ is $CP5$. We define these actions as policies. Hence, one of the policies in Fig. \ref{fig:f6}(c) is $<CP1,CP3,CP5>$. However, this policy has a transition probability, i.e. $<(CP1: 0.7, CP3: 0.2, CP5: 0.5)>$. According to \cite{chen2013combinatorial}, the optimal policy generates the largest combined transition probabilities among all the policies.

We implement the MDP as the Q-learning process \cite{watkins1992q}. Note that other deep learning approaches could be implemented in the reduced topology approach \cite{bertsekas2018feature,9049451}. \textit{However, we do not focus on providing a comparative study of machine learning approaches in this paper, but rather we use a sound existing approach to solve our concerned problem.} 
We evaluate the effectiveness of Q-learning in bootstrapping reputation using CP-nets. In future work, we will compare the performance of the proposed approach with other deep reinforcement learning approaches.

The Q-learning algorithm is a simple reinforcement learning algorithm that learns the value of the state-action sequences through experiences or an iterative process. Those values are stored in a matrix called $Q[S^{A},CP_{A}]$. Initially, the matrix is filled with 0 values. Through an iterative process, the matrix gets filled up with different state-action values. A higher value in $Q[S^{A}, CP_{A}]$ represents that the probability to get rewarded by choosing the CP-nets, $CP_{A}$ for the abstract service, $S^{A}$ is very high. Here, the reward is the closeness of the fractional composite reputation to the optimal reputation. If $OCR$ is the optimal composite reputation and $CCR$ is the current composite reputation after selecting a CP-nets, i.e., the action $a$ for an abstract service, i.e, the state $s$, the reward ($r$) for choosing $CP_{A}$ for $S^{A}$ can be expressed as follows:
\vspace{-3mm}
\begin{equation}
\small
\label{eq:reward}
r[S^{A},CP_{A}] = \frac{1}{| OCR - CCR |} 
\end{equation}     

The value in $Q[S^{A},CP_{A}]$ describes the importance to take the current selection to find the optimal composition. $Q[S^{A},CP_{A}]$ indicates that we have to consider  the current fractional reputation as well as the future aggregated reputation that can be generated from other abstract services and the selection of corresponding CP-nets. The Equation \ref{eq:q4} describes the update process. First, it calculates the current reward or reputation value using Equation \ref{eq:reward} and then checks all the possible other abstract services in the topology and their corresponding actions. The abstract service that has the maximum future reward is chosen for the next state. The current state is updated based on the next state's expected reward or bootstrapped reputation value. In Equation \ref{eq:reward}, $\alpha$ is the learning rate and $\gamma$ is the discounted factor. The learning rate $\alpha$ decides the speed of convergence. $\gamma$ decides weights of the current reward against the future reward in the update process. Usually, $\gamma$ is 0.5, so both rewards are given equal importance in the update process. The process is terminated when no updates on Q values is possible (convergence values) (Algorithm \ref{alg:qlearning}).     
\vspace{-1mm}
\begin{align}
\small
\label{eq:q4}
Q[S^{A},CP_{A}] = (1-\alpha)&Q[S^{A},CP_{A}] + \alpha (r[S^{A},CP_{A}]\\ \notag
&+\gamma max_{\acute{CP_{A}}}Q[\acute{S^{A}},\acute{CP_{A}}]) 
\end{align}  
   
\begin{algorithm}
\fontsize{8pt}{8pt}\selectfont
    \caption{The Q-learning process for the CP-nets composition in bootstrapping reputation}
    \label{alg:qlearning}
    \begin{algorithmic}[1]
    \STATE Initialize $Q[S^{A},CP_{A}]$ to 0
    \STATE $\alpha$ $\gets$ A user defined learning rate in [0,1] 
    \STATE $\gamma$ $\gets$ A user defined discounted factor in [0,1] 
    \FOR {each episode}
    \STATE $S^{A}$ $\gets$ the starting state $s_{0}$
    \STATE number of visited abstract in a topology $o\gets 1$
        \WHILE{$o \neq$  total number of abstract services}
             
             \STATE Choose a CP-nets $CP_{a}$ for $S^{A}$ using $\epsilon$-greedy policy
             \STATE  Calculate current reputation $r[S^{A},CP_{A}]$ 
             \STATE observe the possible next abstract services $\acute{S^{A}}$
             \STATE $Q[S^{A},CP_{A}] = (1-\alpha)Q[S^{A},CP_{A}] + \alpha (r[S^{A},CP_{A}]$
             \STATE $Q[S^{A},CP_{A}] = Q[S^{A},CP_{A}]+\gamma max_{\acute{CP_{A}}}Q[\acute{S^{A}},\acute{CP_{A}}]) $
             \STATE  $o \gets o+1$ 
        \ENDWHILE
    \ENDFOR
    \end{algorithmic}
\end{algorithm} 
    
The learned $Q[S^{A},CP_{A}]$ is used to produce the optimal composition using Equation \ref{eq:q5} \cite{sajibbook2018}. The optimal composition is found in $\epsilon$-greedy fashion. The $\epsilon$-greedy approach traces the best CP-nets following the links with the highest Q-values at each abstract service or state.
\begin{align}
\label{eq:q5}
\pi^{*}(s) = & argmax_{CP_{A}}Q[S^{A},CP_{A}] \\ \notag
& \text{where } \forall  \;CP_{A} \in \text{set of candidate CP-nets} 
\end{align}
 
\vspace{-5mm}
\section{Experimental Results}
We conduct a set of experiments to evaluate the proposed reputation bootstrapping approach using real-world data sets. First, we evaluate the efficiency of the \textit{random approach}, \textit{domain-best approach}, \textit{brute force approach}, and \textit{reduced topology approach} using reputation interdependence in different environmental settings. Next, we compare the accuracy, run-time efficiency, scalability of the  proposed approach to six state-of-the art methods: a) \textit{layer-based Random Forest (LRF)} \cite{lieicsoc2017}, b) \textit{Inheritance Mechanism (IM)} \cite{DBLP:conf/wecwis/NguyenYZ12}, c) \textit{Linear Aggregation (LA)} \cite{ramirez2017evolutionary} (i.e., composition rules without reputation interdependence), d) \textit{Linear Regression (LR)} \cite{Borges:2016:PPG:2972958.2972966}, e) \textit{LSTM-based  matrix  factorization (PLMF)} \cite{8456329}, and f) \textit{DNN approach} \cite{silver2016mastering}. 

The layer-based approach applies a data-driven approach, i.e., random forest to determine the importance of reputation-related factors for a new single or atomic service \cite{lieicsoc2017}. The inheritance mechanism applies weighted aggregation on the past reputations of the existing services of a provider to bootstrap the reputation of the new single service \cite{DBLP:conf/wecwis/NguyenYZ12}. 
The LA approach \cite{ramirez2017evolutionary} does not consider the interdependence among component services. The QoS of the composition is derived using the average or the weighted average of QoS of each component services in the LA approach. Such linear aggregation technique is widely used in web service reputation bootstrapping approaches \cite{DBLP:conf/atal/BurnettNS10,DBLP:conf/trustcom/JiaoLLL11, DBLP:journals/vldb/MalikB09}. The LR approach \cite{Borges:2016:PPG:2972958.2972966,lieicsoc2017} is a supervised machine learning approach where the target reputation value is predicated using independent reputation-related factors. As the independent and dependent variables or reputation-related factors are labelled and their historical data could be represented in a feature matrix, deep learning based approaches are intuitively fit to predict the composite reputation. We consider the PLMF approach which is a LSTM-based matrix factorization approach that captures the dynamic latent representations of QoS parameters in historical data \cite{8456329}. We also consider the Deep Neural Network (DNN) which is a class of machine learning algorithms that uses multiple layers to progressively extract higher level features from the training dataset \cite{silver2016mastering}. The reputation bootstrapping accuracy of these different approaches are compared against \textit{the ground truth}, i.e., reputation values of Github repositories. All the experiments are conducted on computers with Intel Core i7 CPU (8 Cores, 2.30 GHz, 64GB RAM, and RTX 2080 GPU) and programmed in Python. 
\vspace{-3mm}
\subsection{Experimental Setup}
Github is a widely used data set for a range of applications \cite{beller2017travistorrent}. Because of the modeling and behavior similarities to our research focus, we selected Github dataset as the appropriate platform to evaluate our proposed approach. We consider public repositories in Github as composite services. The \textit{insight} of a repository specifies both the dependent and dependencies or submodules which are publicly available. For example, the insight of the repository `mopidy/mopidy'\footnote{https://github.com/mopidy/mopidy} depicts that the repository has 237 watchers, 261 dependent repositories, 13 dependencies, 106 contributors, its community or owner has 22 other repositories with average 1000 ratings or stars and 120 followers. The Github also provides insights about the commits, code frequency, and forks of the public repositories.

We use Github Developer REST API \cite{githubDev} to retrieve the public repositories. A repository may contain submodules \cite{githubDev}.  Submodules allow an owner to keep a Git repository as a subdirectory of another Git repository. This lets the owner to clone another repository into their projects and to keep their commits separate. Hence, submodules, i.e., repositories from other projects are considered as the component services and the parent repository is considered as the composite service. We collected 300 repositories in the $Set \;A$, where each repository has \textit{4 or more submodules} using a Github web crawler, `Scrapy'\footnote{https://github.com/scrapy/scrapy}. 
Each composition topology in Github is treated as a new composition in the proposed Q-learning approach. The number of reputation-related factors varies from 4 to 13 in the compositions of the Github dataset. The Github dataset also provides the history of invocations (pull requests) among component services. As the proposed approach uses a generic model to leverage the reputation-related factors, the diversity of reputation-related factors in the Github dataset is a perfect match to test the scalability of the proposed approach. Experiments in different domains would provide a more robust evaluation of the proposed approach. However, finding the right data set proved to be quite challenging as to the best of our knowledge, we could not find any public datasets other than Github that have reputation-related factors, component services' ratings, composition topologies and invocation history.
\begin{figure*}[t!]
   \centering
   \vspace{-10mm}
   \subfloat[]{\includegraphics[width=0.4\textwidth, height=0.3\textwidth]{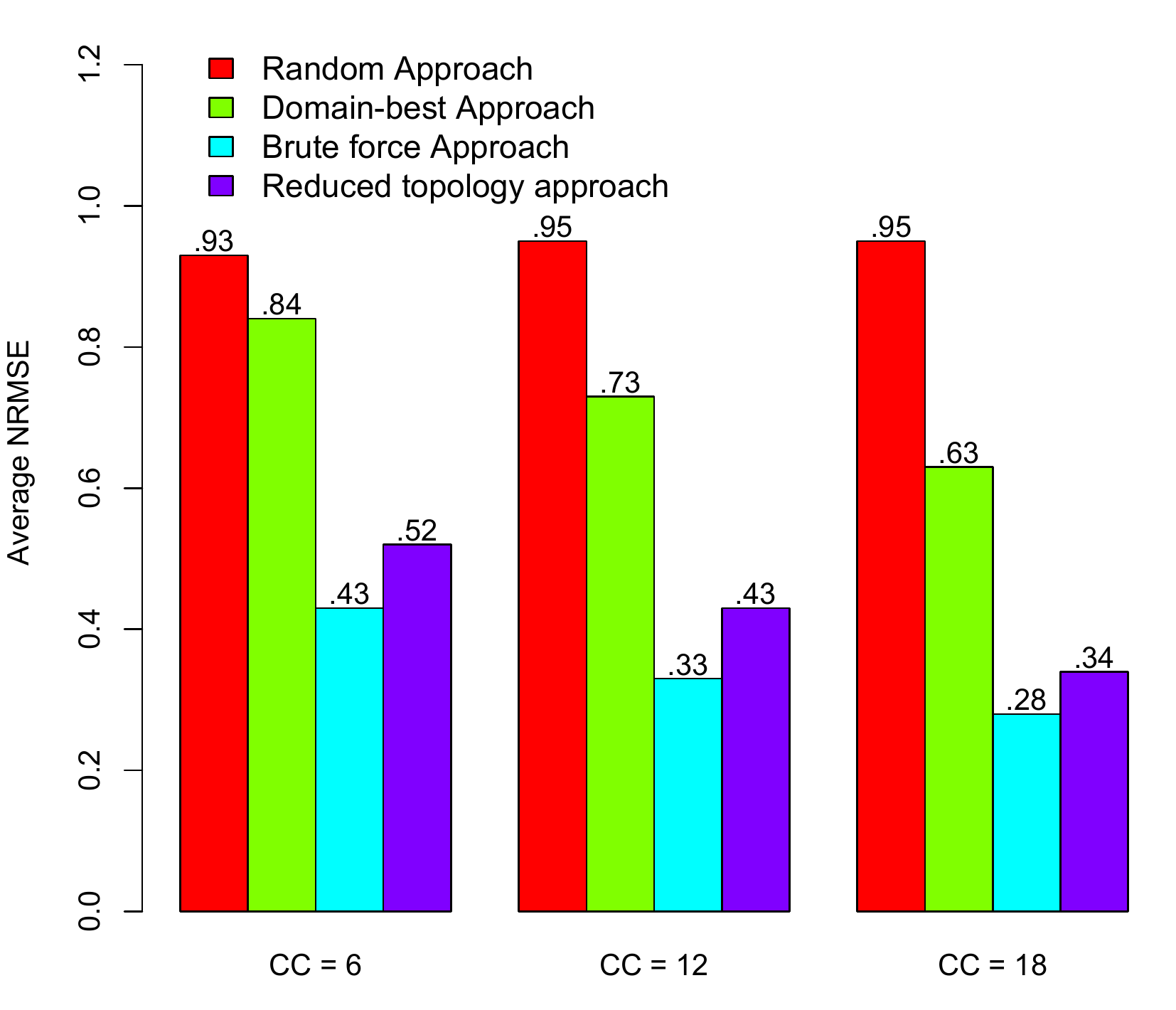} } 
      \subfloat[]{\includegraphics[width=0.33\textwidth, height=0.3\textwidth]{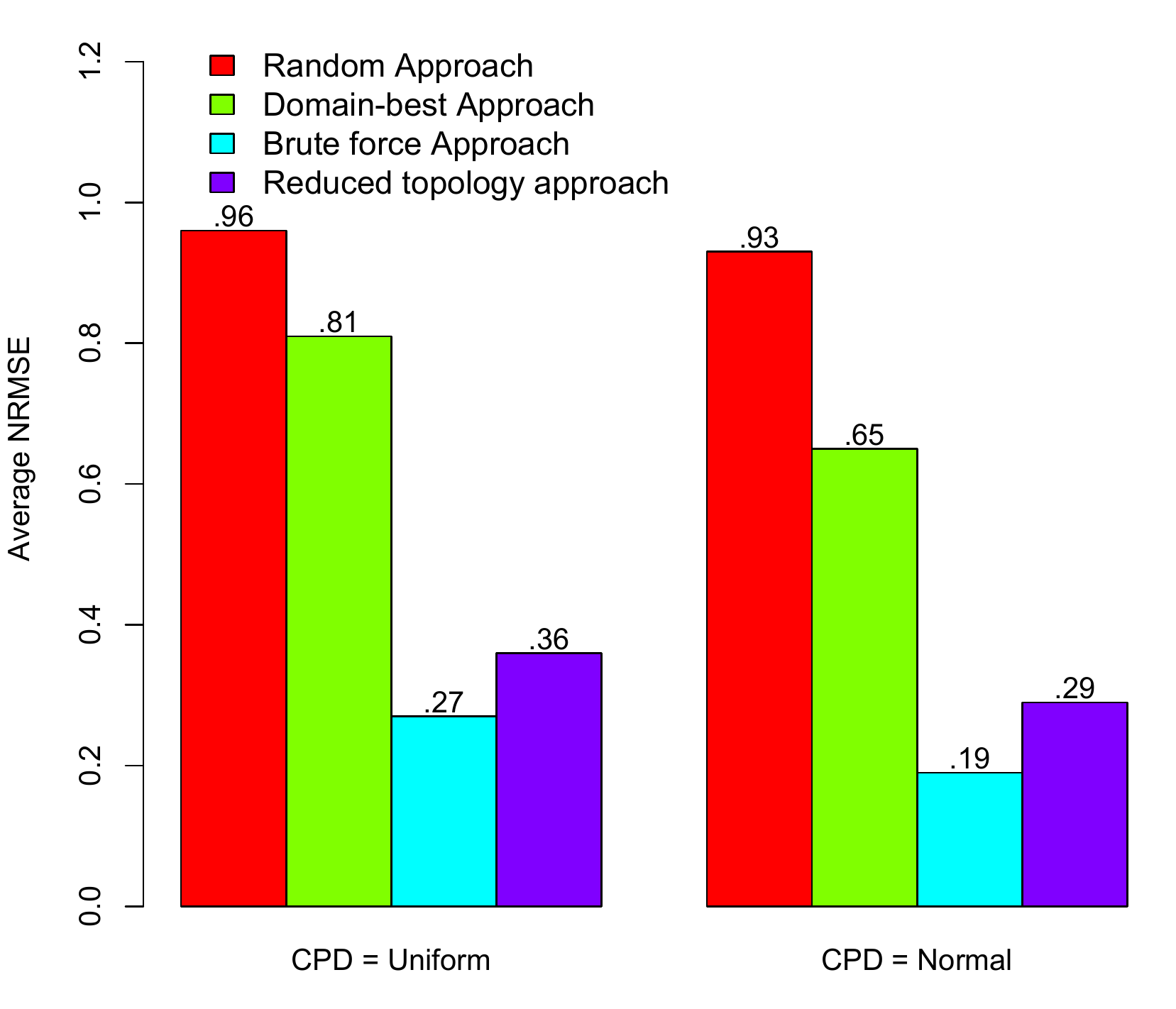}}  \subfloat[]{\includegraphics[width=0.26\textwidth, height=0.3\textwidth]{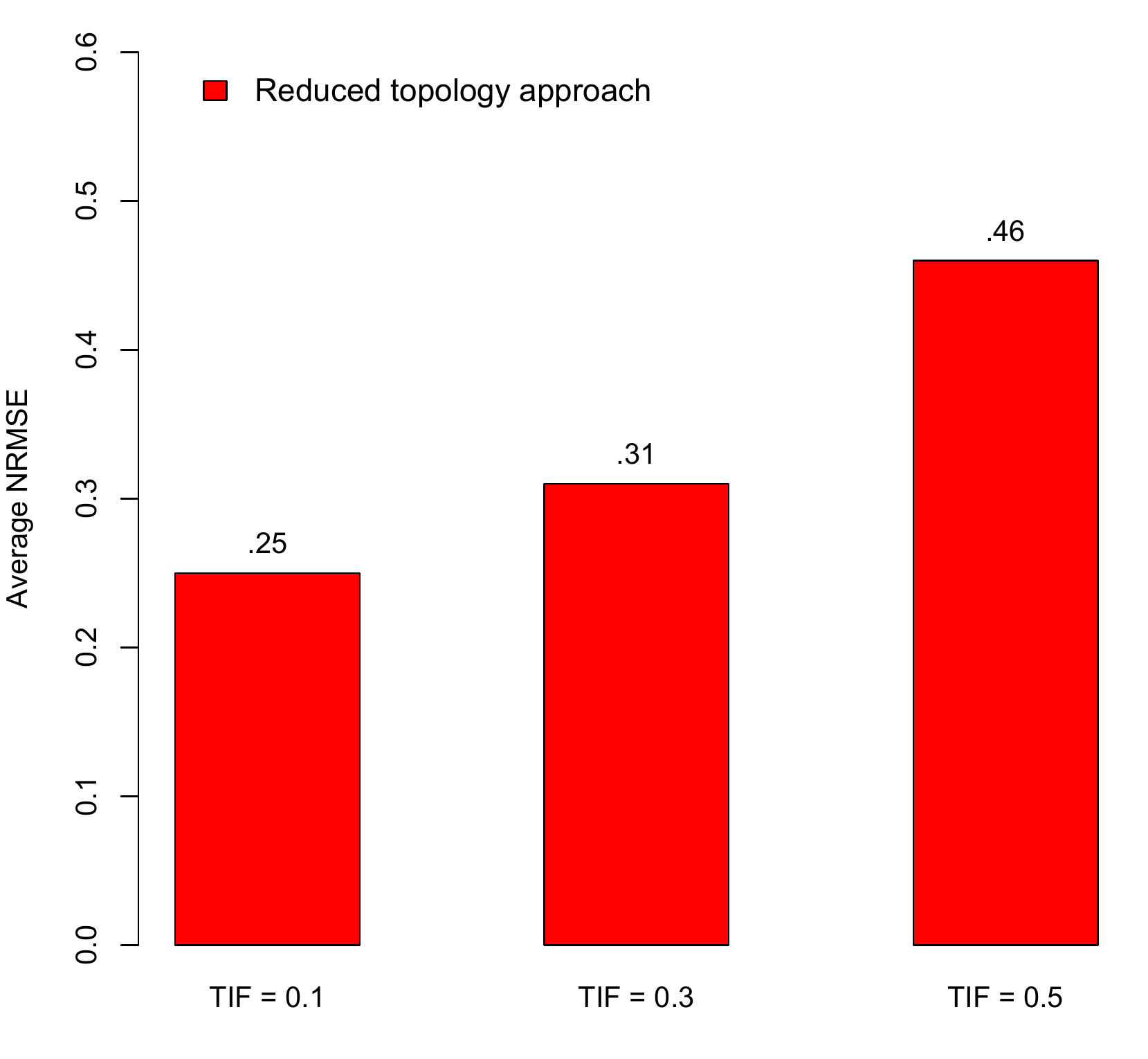}} \\
      \vspace{-2mm}
  \caption{\small Effects of (a) Candidate CP-nets (\textit{CC}), (b) CP-nets mapping distribution (\textit{CPD}), and (c) Threshold of Influence Factor (\textit{TIF}) \normalsize}
   \label{fig:5}
   \vspace{-5mm}
 \end{figure*} 
 


The proposed framework has a set of environment parameters that need to be set before its application. Such environment variables are the number of candidate CP-nets, CP-nets mapping distribution, Composition topology, service invocations and the threshold value of Influence Factor to reduce the topology. The efficiency of the proposed framework depends on the values of the environment variables. Table \ref{tab:set} describes the environmental variables used in the experiments. The candidate CP-nets ($CC$) are generated with randomly picked 4 reputation-related factors. Maximum $4! = 24$ candidate CP-nets could be generated for each component service. The parameter, $CC: (6, 12, 18)$ covers the effective range, i.e., closer to the minimum, median and closer to the maximum value. The lowest value in the Threshold of Influence Factor (TIF), i.e., $TIF = 0$ implies that component services are not reduced from the topology. A higher value, i.e., $TIF = 0.5$ implies that 50\% of component services are reduced from the topology which is a large reduction in practice. The parameter, $TIF: (0.1, 0.3, 0.5)$ covers the effective range, i.e., closer to the minimum, median and closer to a practical large value.  

\begin{table}
  \caption{Simulation environment settings}
  \vspace{-3mm}
  \label{tab:set} 
\begin{center}
\begin{tabular}{ |p{4cm}|p{4cm}| } 
 \hline
 \textbf{Variables}&  \textbf{Parameters}\\ 
 \hline
 Reputation of a provider & Average stars of the contributors \\ 
 \hline
 Reputation of a community &  Average stars of the owner\\ 
 \hline
 Reputation of Interrelated Services & Apply Equation \ref{eq:s1}\\
  \hline
 Number of candidate CP-nets ($CC$) &  6, 12, 18 \\ 
 \hline
 CP-nets mapping distribution ($CPD$) & Uniform, Normal  \\ 
 \hline
 Maximum Number of Component Services ($N$)& 30 \\ 
 \hline
 Highest depth of dependency tree ($DT$)& 20 \\ 
 \hline
 Service Invocation ($SI$)& Number of common watchers between repositories\\ 
 \hline
 Threshold of Influence Factor ($TIF$) & 0.1,0.3,0.5  \\ 
 \hline
 Reinforcement learning rate ($\alpha$) & 0.2,0.5,.08  \\ 
 \hline
 Discounted Reward rate ($\gamma$) & 0.5  \\ 
 \hline
\end{tabular}
\end{center}
\vspace{-5mm}
\end{table}

The values in CPTs are semantically mapped into 5 segments, i.e., \textit{Very High}, \textit{High}, \textit{Moderate}, \textit{Low}, and \textit{Very Low}. For a provider, the highest number of stars of a contributor is the upper bound and the lowest number of stars of a contributor is the lower bound. Similarly, for a provider, the highest number of stars of an owner is the upper bound and the lowest number of stars of an owner is the lower bound. We use standard values for the Q-learning approach, i.e., $\alpha = 0.5$ and $\gamma = 0.5$. \textit{Note that finding the optimal learning and discounted reward rate is out of the focus of this paper.}
\vspace{-2mm}
\subsubsection{Implementing LA approach}
The LA approach is a generic approach, hence we need to modify it to focus on a specific QoS, i.e., reputation. First, we represent the topology independence as follows:

\begin{itemize}[leftmargin=*, nolistsep]
\item The number of incoming and outgoing edges from a component service is the same. It represents that a component service influences all other services in a similar manner and gets influenced by all other services equivalently. 
\item The weights of the edges have the same small value. It is represented as $\theta$ and  $\theta > 0$. $\theta$ represents that the influence value is small and as all are influencing similarly, they nullify each other's influence. Hence, the topology is independent of reputation influence. 
\end{itemize} 


We calculate the reputation influence factor of an independent component service using Equation \ref{norif}:
\vspace{-2mm}
\begin{align}
\label{norif}
\small
&AOW = \frac{n \times \theta }{n} = \theta,
IW = max(\theta, \theta,.....,\theta) = \theta \\\notag
&\text{If } AOW = IW, RIF = \frac{IW}{AOW} =  \frac{\theta}{\theta} = 1
\end{align} 

We apply the Random Forest (RF) to determine the relative importance of reputation related factors for each component service \cite{DBLP:conf/atal/BurnettNS10}. The corresponding CP-nets, $CP_{ij}$ of the component service, $S^{C}_{ij}$ is constructed following the ranking of influence, i.e., the direction of the conditional graph is from the influential node to the least influential node. $RR(CP_{ij}, S^{C}_{ij})$ is the mean reputation value using equation \ref{eq:repucpcal}. Hence, the reputation of the composite service without the reputation interdependence ($\overline{CR}$) is calculated using the aggregated reputation of each component service without their weighted influence factor (as $RIF = 1$). The composition rules without the reputation interdependence are derived from Equation \ref{eq:Rcal} as follows:
\vspace{-2mm}
\begin{equation}
\small
\label{eq:Rcalwithout}
\overline{CR} = \frac{RR(CP_{ij}, S^{C}_{ij})}{\sum RR(CP_{ij}, S^{C}_{ij})} 
\end{equation}
\subsubsection{Implementing LR approach}
In our dataset, the reputation ground truth of compositions is the target value and the reputation-related factors of each component services are independent variables. We use scikit learn python library\footnote{https://scikit-learn.org/sklearn.linear\_model.LinearRegression.html} to import the linear regression model. We fit the model on the training data (70\% of total rows in the dataset) and predict the values for the testing data (30\% of total rows in the dataset). We use R2 score\footnote{https://scikit-learn.org/sklearn.metrics.r2\_score.html} to measure the accuracy of our model.
\vspace{-3mm}
\subsubsection{Implementing PLMF approach}
We collect 300 composite services which includes total 7555710 repositories and 10 reputation indicators for each repository. The matrix [7555710 X 10] acts as the input data set for the PLMF \cite{8456329}. The training set has different data density from 5\% to 20\% with a step size 5\%. We apply double Layer LSTM (gated recurrent neural network) with 108 hidden neurons in the first layer, 54 hidden neurons in the second layer, and dropouts=0.2 between the hidden layers and the output layer. The prediction accuracy of the four different matrix densities, i.e., 5\%, 10\%, 15\%, and 20\% are averaged as the output of the PLMF approach. 
\vspace{-3mm}
\subsubsection{Implementing DNN approach}
We implement the standard DNN \cite{silver2016mastering} for prediction which consist of three layers. The input layer is constructed by the types of reputation indicators: a) provider (PI), b) community (CI), c) Similar Service (SSI), and d) Insight (II). The output layer consists of the reputation or start values (ground truth). The DNN is implemented in Python with packages Scikit-learn and Tensorflow. Initially, the default parameters of these implementations are chosen in the experiments. Later, these values are tuned using a grid search with the training dataset. We apply K-fold cross validation [21] where ($k$ = 5), so there are 5 iterations of model training and testing. Each iteration contains different sets of 80\% data for training, and 20\% data for testing.

\vspace{-3mm}
\subsection{Accuracy and Runtime Efficiency}
We use the number of stars ($GS$) of a repository as the \textbf{ground truth}. The proposed approach represents reputation in the [0,1]. We transform the reputation $r_{i}$ into number of stars using the number of stars in its submodules as follows:
\begin{align}
\small
\alpha = max(&\text{highest number of stars of a contributor,} \\ \notag &\text{number of stars of a owner})  \\\notag
\text{Predicted} &\text{ number of stars, } PS = r_{i} \times \alpha
\end{align}

The proposed framework will be efficient if $PS$ becomes close to $GS$ for a repository. As we have $m = 300$ repositories as composite services, we generate 300 stars as the predicted stars from the submodules or component services. The efficiency of the proposed approach is calculated using the Normalized Root Mean Square Error (NRMSE) in Equation \ref{eq:rmse}. A lesser value of NRMSE imposes a better efficiency of the proposed approach. 
\begin{align}
\label{eq:rmse}
\small
\small NRMSE = &\sqrt{\frac{\sum_{i=1}^m(1- \frac{PS_{i}}{GS_{i}})^2}{m}} \normalsize
\end{align}


We compare the efficiency of the \textit{random approach}, \textit{domain-best approach}, \textit{brute force approach} and \textit{reduced topology approach} using reputation interdependence (Equation \ref{eq:Rcal}). We consider all the combinations of environment variables. For example $(CC: 6, CPD: Uniform, TIF:0.1)$ is a configuration. First, we discuss the effect of the number of candidate CP-nets ($CC$) in the proposed approaches. According to Fig. \ref{fig:5}(a), the NRMSE reduces for all approaches except the random approach when more candidate CP-nets are considered. The decreasing rate is almost similar to the other three approaches. The possible reason for this behavior is that a larger number of candidate CP-nets holds a wider spectrum of relative importance. Hence, the probability of selecting the optimal CP-nets is higher in the composition. One key issue of using a large candidate number is that it may increase the runtime computation. 

\begin{figure*}[t!]
   \centering
   \vspace{-10mm}
   \subfloat[]{\includegraphics[width=0.33\textwidth, height=0.3\textwidth]{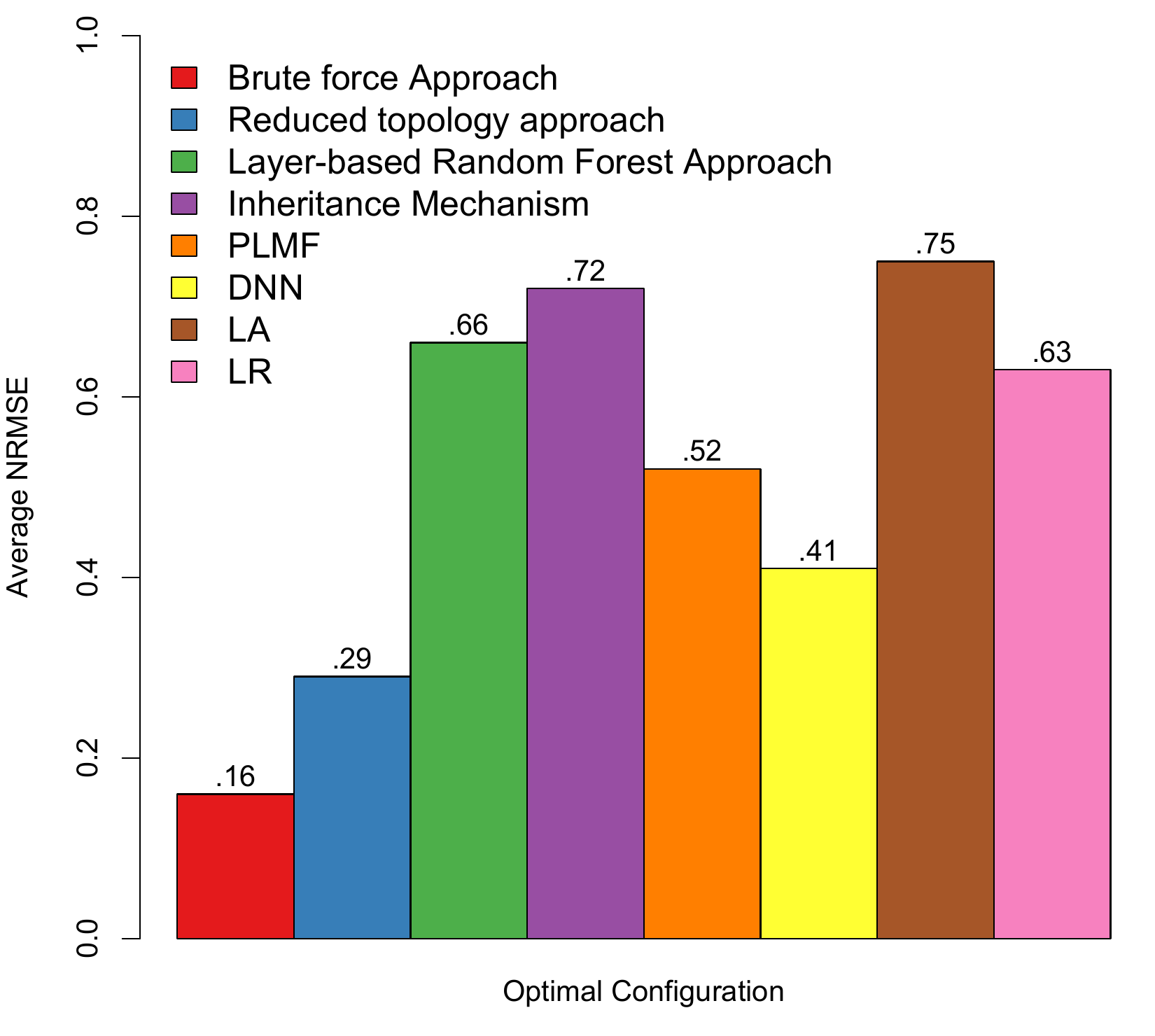}} 
      \subfloat[]{\includegraphics[width=0.33\textwidth, height=0.3\textwidth]{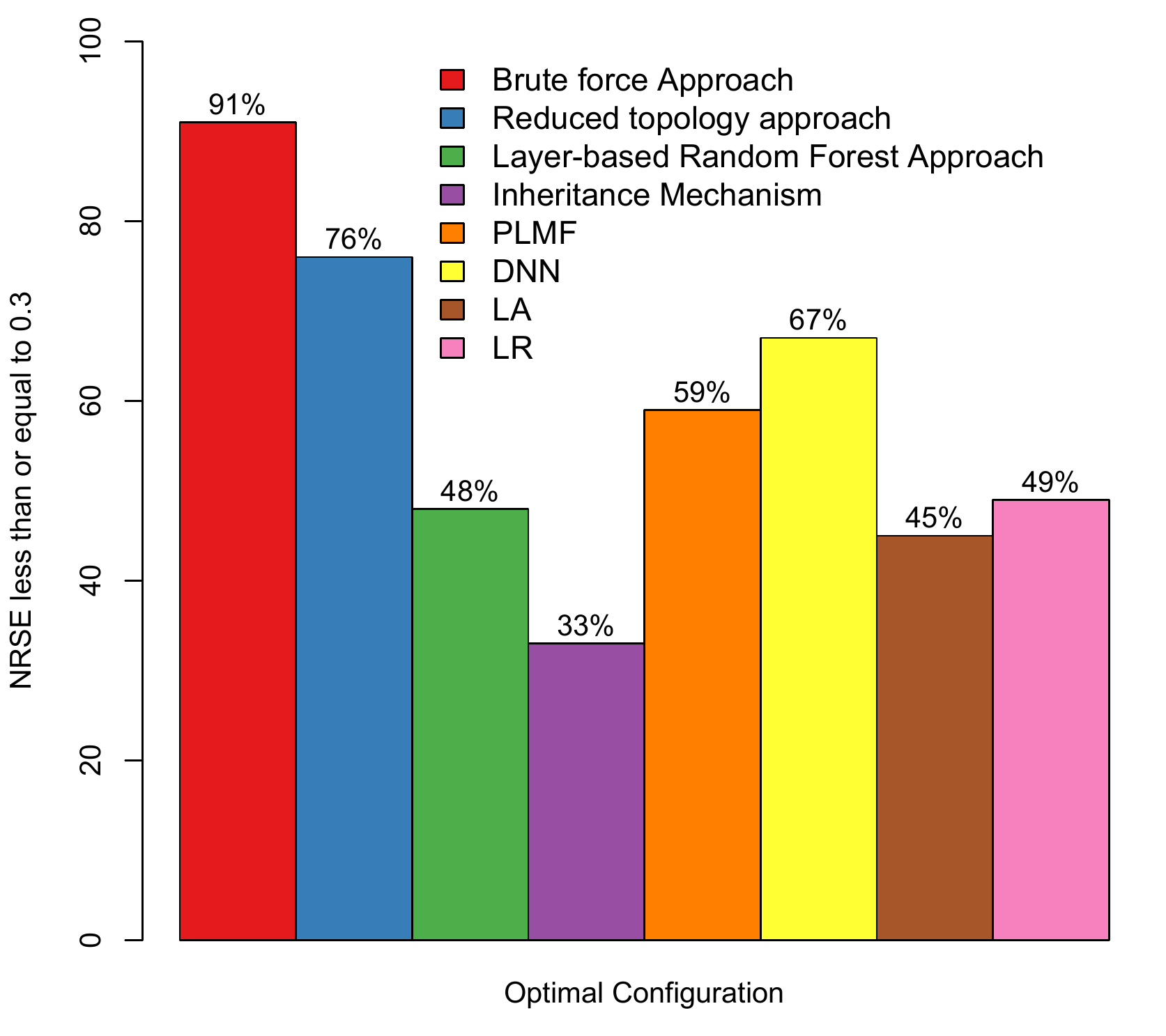}}  \subfloat[]{\includegraphics[width=0.33\textwidth, height=0.3\textwidth]{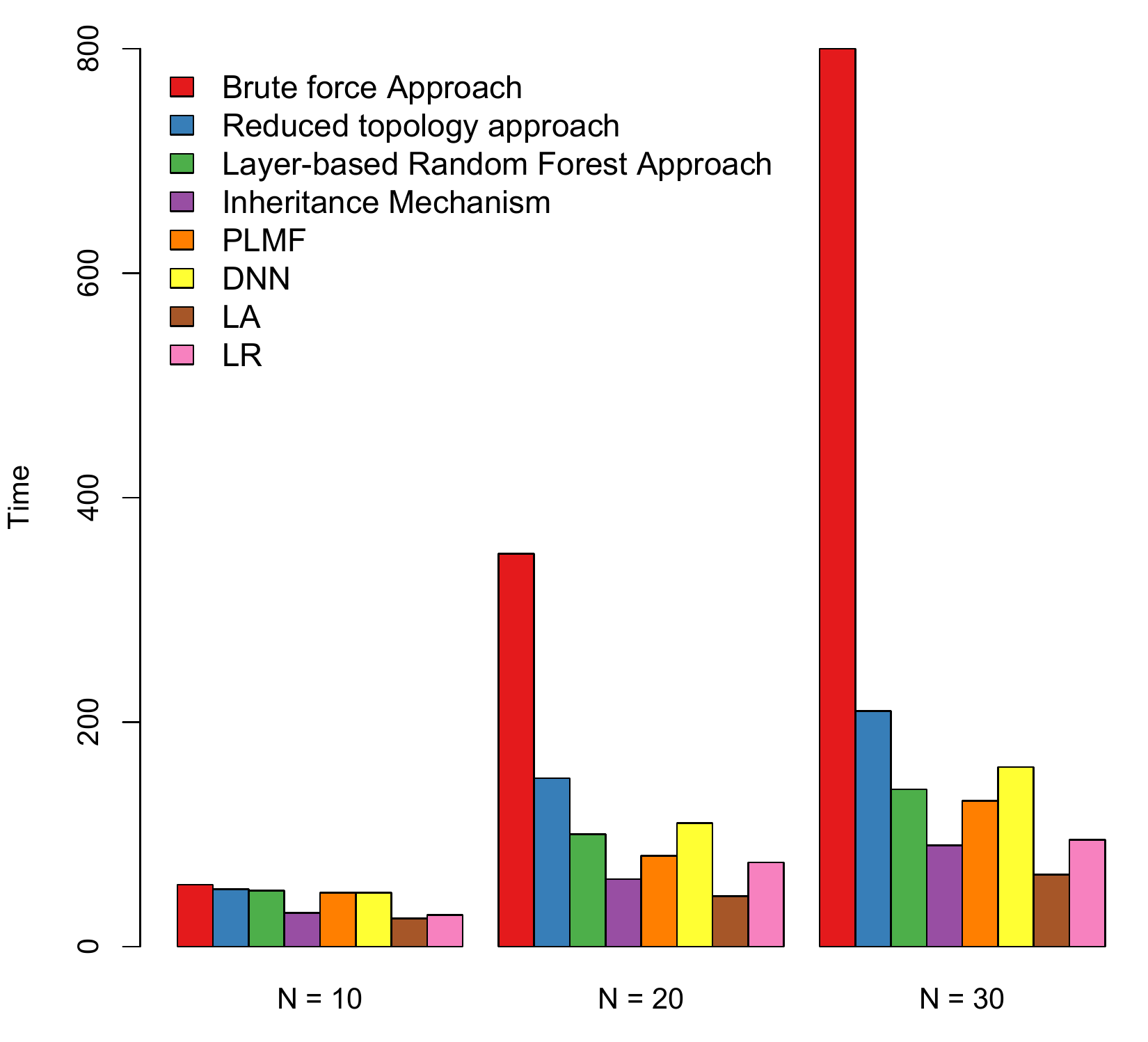}} \\
      \vspace{-2mm}
   \caption{(a) Average bootstrapping error, (b) Acceptance ratio, and (c) Runtime complexity \vspace{-5mm}}
   \label{fig:6}
 \end{figure*}  

Next, we discuss the effect of CP-nets mapping distribution ($CPD$) in the proposed approaches. According to Fig. \ref{fig:5}(b), the NRMSE reduces for all approaches except the random approach when the Normal distribution is used. The possible explanation is that relative importance in CP-nets is not uniform in nature. Next, we analyse the effect of the Threshold of the Influence Factor (TIF) in Fig. \ref{fig:5}(c). Note that the TIF is only applicable to the proposed reduced topology approach as it is used as a heuristic to reduce the topology. As a result, Fig. \ref{fig:5}(c) only shows the effect of the reduced topology approach with different TIF values. According to Fig. \ref{fig:5}(c), a lower $TIF$ value produces lower average NRMSE than higher $TIF$ value. It means that the reduced topology approach performs similarly to the brute force approach when $TIF$ is significantly low. However, the runtime of a lower $TIF$ may not be better than the brute force approach. Hence, a moderate $TIF$ value is appropriate for the topology reduction approach. Next, we run the proposed approach in 18 different simulation settings (generated from the different combinations of $CC$, $CPD$, and $TiF$). Finally, we find an optimal configuration, i.e., $(CC: 12, CPD: Normal, TIF: 0.3)$ where the proposed brute force approach and reduced topology approach produce the lowest average NRMSE. This optimal configuration is used to compare the proposed approach against the six competing approaches.

According to Fig. \ref{fig:6}(a), the brute-force approach produces the lowest NRMSE 0.16 and the reduced topology approach produces the second best average NRMSE 0.29. Its accuracy is significantly higher (around 43\%) than the LRF, IM, LA and LR. Note that LRF, IM, LA and LR produce the similar result. The key reason is that these approaches are primarily designed for atomic services, and do not cluster reputation-related factors based on the topology in a composition. Although DNN and PMF cluster the topologies in their feature factorization process, they do not consider the semantic interpretation of reputation influence and the reputation propagation direction. As a result, compositions with similar topology, but unrelated reputation-related factors may be clustered which may cause over-fitting or under-fitting learning issues. We find that the proposed approach is around 14\% and 21\% higher accurate than the DNN, and PLMF respectively. According to Fig. \ref{fig:6}(b), the brute force approach and the reduced topology approach have the highest probability of 0.91 and 0.76 respectively to produce the average NRMSE 0.3. Hence, the confidence of accuracy in the proposed reduced topology approach is significantly higher than other approaches. According to Fig. \ref{fig:6}(c), the brute-force approach has the exponential runtime. Here, $N$ is the number of abstract services in the composition topology. This approach should not be used in larger compositions. The LRF, IM, LR, LA, PLMF, and DNN approach have relatively higher run-time efficiency (around 8\%) than the proposed approach. The key reason is that proposed approach applies reinforcement learning to compose the CP-nets and has an extra process of transforming reputation values from the composite CP-nets, and vice-versa. 
\vspace{-3mm}
\subsection{Scalability}


\begin{figure*} [t!]
    \centering
    \vspace{-10mm}
      \subfloat[]{\includegraphics[width=0.42\textwidth, height=0.3\textwidth]{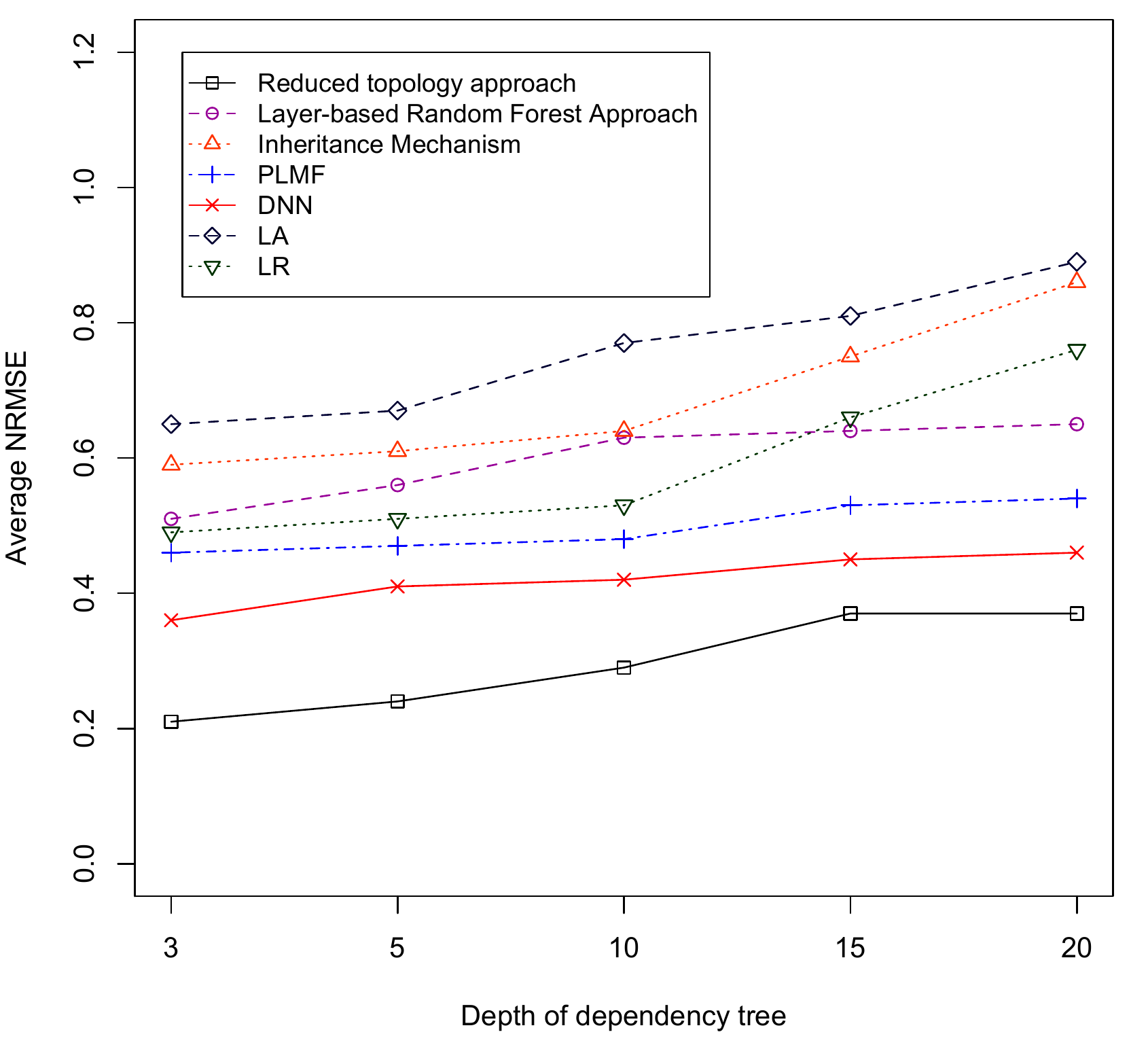}} \hspace*{-.01em} 
      \subfloat[]{\includegraphics[width=0.42\textwidth, height=0.3\textwidth]{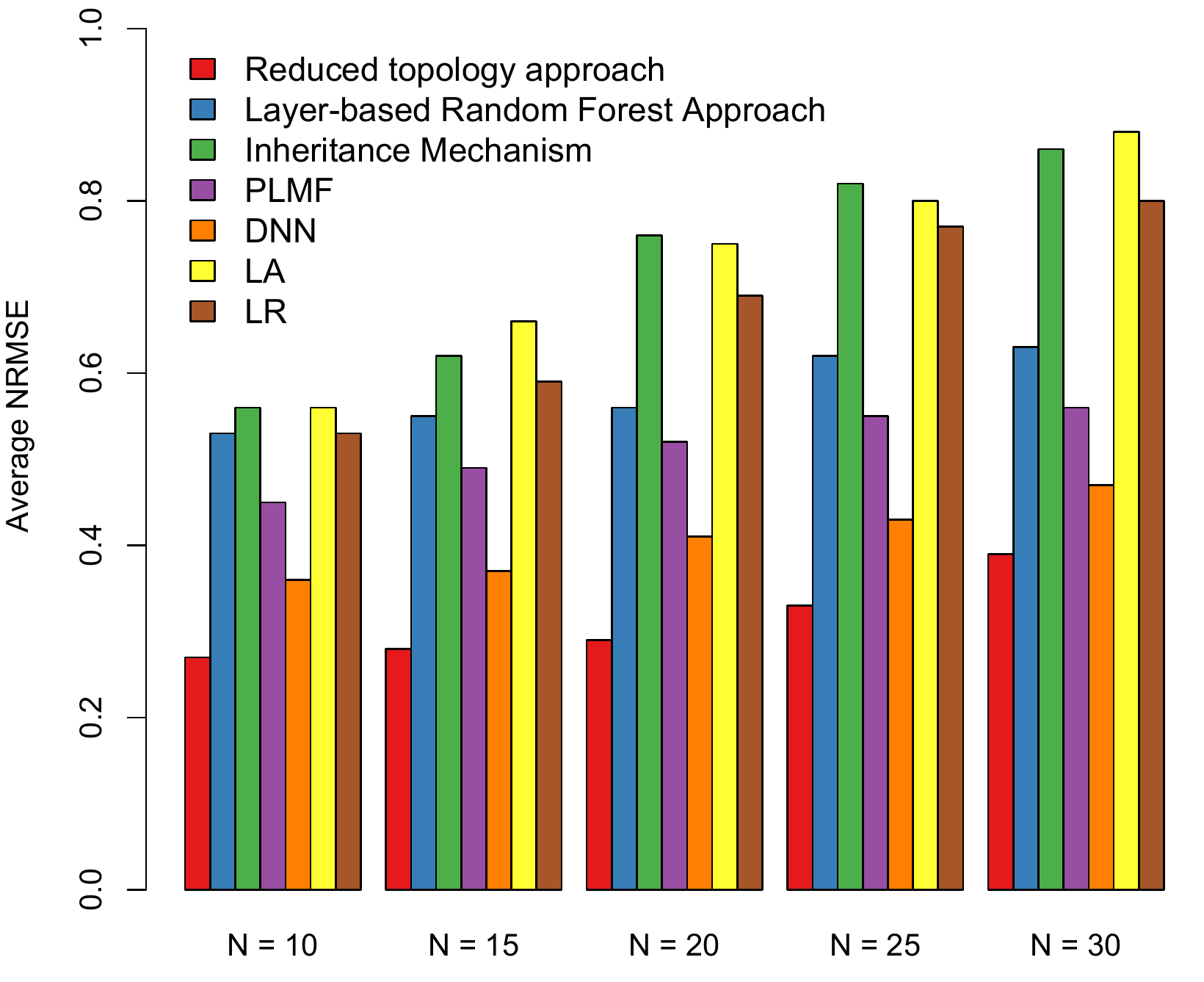}}  \\ 
      \vspace{-2mm}
    \caption{\small Average performance of different approaches in (a) different depths of dependency tree, (b) different topology sizes \normalsize}
    \label{fig:exp3}
    \vspace{-6mm}
\end{figure*}

\begin{figure*}
    \vspace{-5mm}
    \centering
      \subfloat[]{\includegraphics[width=0.44\textwidth, height=0.3\textwidth]{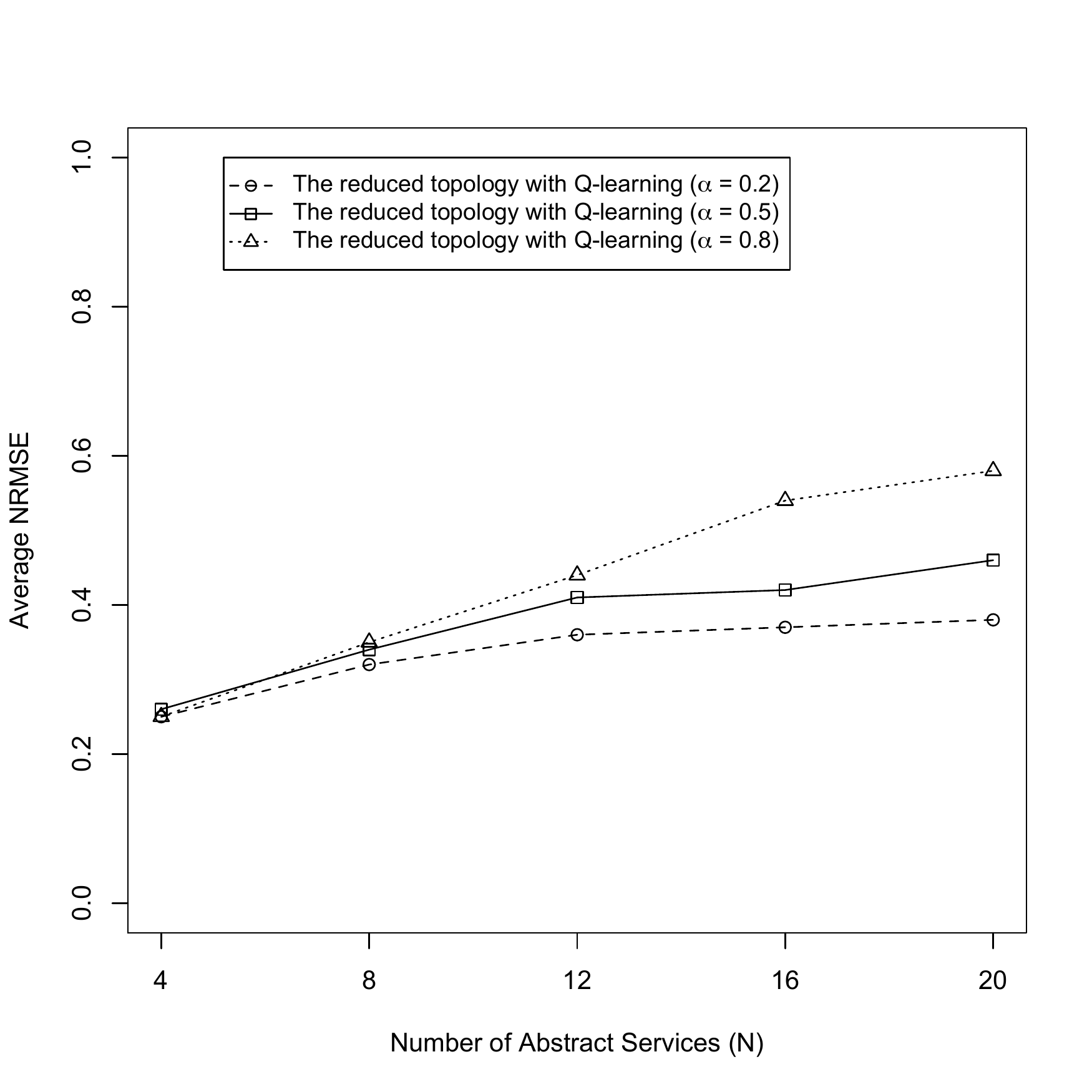}} 
      \subfloat[]{\includegraphics[width=0.44\textwidth, height=0.3\textwidth]{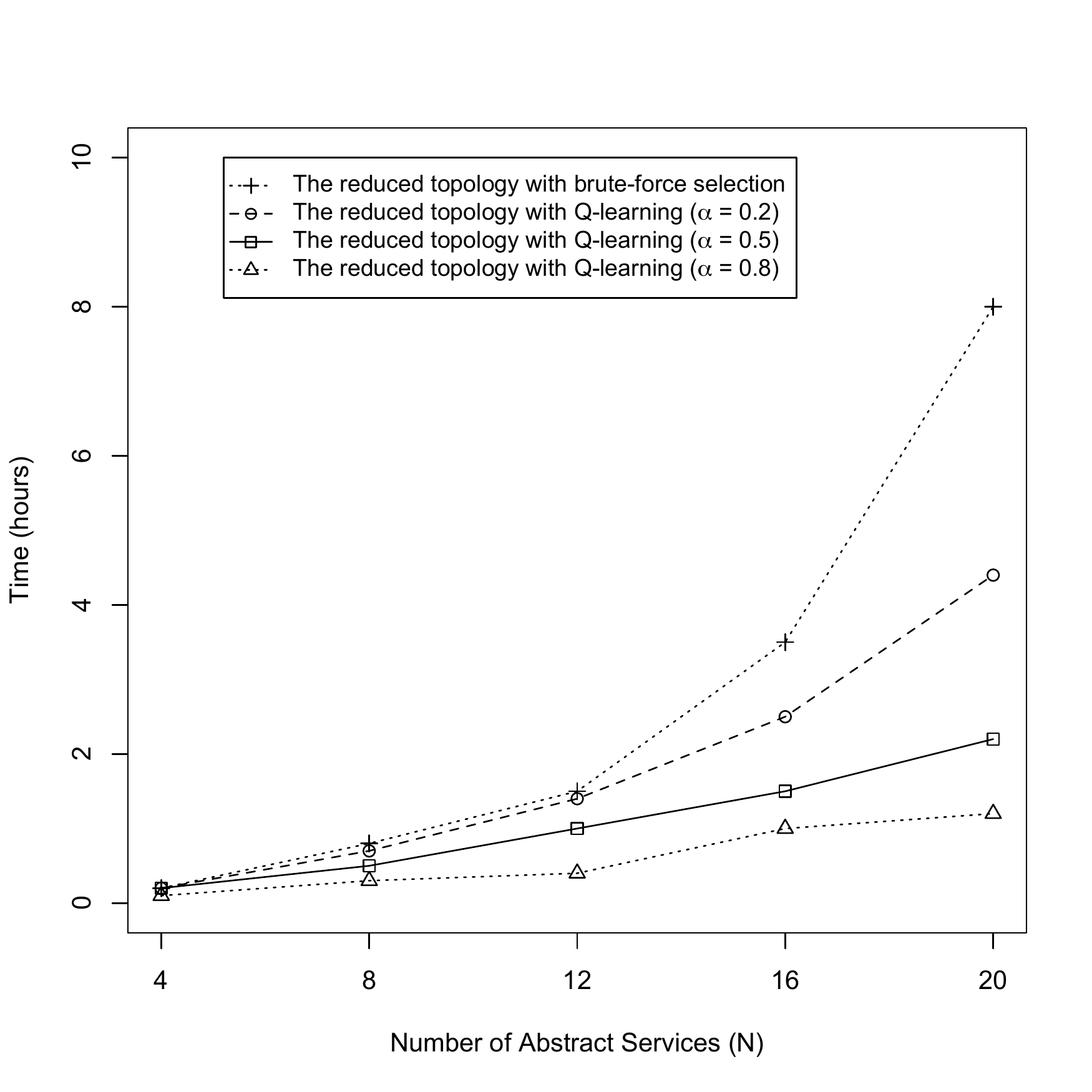}}  \\  
      \vspace{-2mm}
    \caption{\small The proposed reduced topology approach with different learning rates in (a) Accuracy and (b) Runtime efficiency \normalsize}
    \label{fig:lastexp}
    \vspace{-5mm}
\end{figure*}

Fig. \ref{fig:exp3}(a) compares the \textit{scalability} of the proposed reputation interdependence and the heuristic on reduced topology on the result accuracy with the competing approaches. We consider topologies with different depth of dependency trees (3 to 20). A larger depth in the dependency tree specifies a longer linear chain of interdependent component services. We find that the proposed approach, DNN, PLMF, and LRF approaches are scalable where the average standard deviation in mean accuracy is 0.45 for in the wide range of dependencies [3, 20]. of produce an almost similar result with a smaller number of dependencies (i.e., 3 to 10). We also found that the proposed approach is highly scalable as there is only 1.7\% increase in error rates from a short dependency to a long dependency tree. The proposed approach maintains at least 15\% higher accuracy than DNN, PLMF, and LRF in different dependencies. Besides, the proposed approach produces significantly higher accuracy (56\% more) than the LA, LR, and IM approaches for a larger depth of dependency trees. For a larger depth, the LA, LR, and IM approaches tend to give the same weight to every component services and does not consider the cascading effect. Fig. \ref{fig:exp3}(b) compares the efficiency of the proposed approach with the competing approaches in respect of the size of the topology, i.e., the number of component services. We use different sizes (from 10 to 30 component service) of compositions. We find that the all approaches perform close to smaller compositions ($N=10$). However, only the proposed approach produces higher accurate results consistently across different sizes of composition topologies. The LA, LR, and IM approaches are not suitable for larger compositions ($N>15$) as they do not consider the performance correlations among component services. The proposed approach produces at least 33\% higher accuracy than DNN, PLMF, and LRF for larger compositions ($N \geq 20$). \textit{we conclude that proposed reduced topology approach which considers the reputation interdependence and composition topology is the most suitable approach for reputation bootstrapping in both small and large compositions.}    


\vspace{-3mm}
\subsection{The effect of the learning rate}
Finally, we analyze the effect of the learning rate in the proposed reduced topology approach. We use 3 different learning rates: a) low ($\alpha$ = 0.2), b) moderate ($\alpha$ = 0.5), and c) high ($\alpha$ = 0.8) to learn Q-matrices in Algorithm \ref{alg:qlearning}. Figure \ref{fig:lastexp} (a) depicts the accuracy of the proposed approach in different learning rates. It shows that better accuracy is achieved when the learning rate is relatively lower. The accuracy decreases when the learning rate increases.    We find around 35\% decrease in accuracy when higher learning rates are used than the moderate learning rates for large compositions (number of abstract services is more than 15). The key reason is that the Q-learning gives more weights on future transitions and overestimates the outcome in higher learning rates. We find the highest accuracy when the learning rate is the lowest. Figure \ref{fig:lastexp} (b) depicts the runtime efficiency of the proposed approach in different learning rates. The brute-force approach has exponential runtime efficiency as it considers all the possible combinations of candidate CP-nets. We find that a lower learning rate almost behaves as a brute-force approach. The key reason is that the Q-learning gives more weights on current states and explores most of the adjacent states to estimate the outcome in lower learning rates. Due to the higher number of explorations, lower learning rates in the proposed approach provides poor runtime efficiency. The runtime efficiency significantly increases when the learning rate increases. As there is a tradeoff between the runtime efficiency and the accuracy for the selection of learning rates, we find that the moderate learning rate provides the most satisfactory outcome.      
\vspace{-4mm}
\section{Conclusion}
We use CP-nets to represent the reputation of a service. The proposed reduced topology approach applies reinforcement learning (Q-learning) to select the optimal candidate CP-nets using the reputation-interdependence among the component services. Experimental results show that the reputation-interdependence and the composition topology are key elements in the reputation bootstrapping of a composition. Experimental results also show that the proposed reduced topology approach is runtime efficient and produces significantly better results than the DNN, PLMF, LA, LR, LRF, and IM approaches. It requires no extra process (e.g., publishing SLA or a trial period) compared to guarantee-based and trail-based approaches. 
One key limitation of the proposed CP-net based approach is that it is applicable for a static composition topology. In the dynamic environment, the composition topology gets updated as well as the reputation-related factors. In future, we will explore different adaptive techniques in the dynamic composition environment. 
\vspace{-4mm}
\section*{Acknowledgement}

This research was partly made possible by DP160103595 and LE180100158 grants from the Australian Research Council. The statements made herein are solely the responsibility of the authors. \vspace{-3mm}
\ifCLASSOPTIONcaptionsoff
  \newpage
\fi




\bibliographystyle{IEEEtran}
\bibliography{sajibbib}

\begin{thebibliography}{10}
\providecommand{\url}[1]{#1}
\csname url@samestyle\endcsname
\providecommand{\newblock}{\relax}
\providecommand{\bibinfo}[2]{#2}
\providecommand{\BIBentrySTDinterwordspacing}{\spaceskip=0pt\relax}
\providecommand{\BIBentryALTinterwordstretchfactor}{4}
\providecommand{\BIBentryALTinterwordspacing}{\spaceskip=\fontdimen2\font plus
\BIBentryALTinterwordstretchfactor\fontdimen3\font minus
  \fontdimen4\font\relax}
\providecommand{\BIBforeignlanguage}[2]{{%
\expandafter\ifx\csname l@#1\endcsname\relax
\typeout{** WARNING: IEEEtran.bst: No hyphenation pattern has been}%
\typeout{** loaded for the language `#1'. Using the pattern for}%
\typeout{** the default language instead.}%
\else
\language=\csname l@#1\endcsname
\fi
#2}}
\providecommand{\BIBdecl}{\relax}
\BIBdecl

\bibitem{Erl:2004:SAF:983556}
T.~Erl, \emph{Service-Oriented Architecture: A Field Guide to Integrating XML
  and Web Services}.\hskip 1em plus 0.5em minus 0.4em\relax New York, USA:
  Prentice Hall, 2004.

\bibitem{DBLP:journals/internet/MalikB09}
Z.~Malik and A.~Bouguettaya, ``Reputation bootstrapping for trust establishment
  among web services,'' \emph{{IEEE} Internet Computing}, vol.~13, no.~1, pp.
  40--47, 2009.

\bibitem{DBLP:conf/icws/BahutairBN20}
M.~Bahutair, A.~Bouguettaya, and A.~G. Neiat, ``Just-in-time memoryless trust
  for crowdsourced iot services,'' in \emph{Proceedings of the ICWS}.\hskip 1em
  plus 0.5em minus 0.4em\relax {IEEE}, 2020, pp. 1--8.

\bibitem{DBLP:conf/trustcom/JiaoLLL11}
H.~Jiao, J.~Liu, J.~Li, and C.~Liu, ``A framework for reputation bootstrapping
  based on reputation utility and game theories,'' in \emph{Proceedings of
  TrustCom}.\hskip 1em plus 0.5em minus 0.4em\relax Changsha: IEEE, 2011, pp.
  344--351.

\bibitem{6461455}
Y.~{Wu}, C.~{Yan}, Z.~{Ding}, G.~{Liu}, P.~{Wang}, C.~{Jiang}, and M.~{Zhou},
  ``A novel method for calculating service reputation,'' \emph{IEEE Trans. on
  Automation Science and Eng.}, vol.~10, no.~3, pp. 634--642, 2013.

\bibitem{4279708}
X.~Liu and A.~Bouguettaya, ``Managing top-down changes in service-oriented
  enterprises,'' in \emph{Proceedings of the ICWS}.\hskip 1em plus 0.5em minus
  0.4em\relax IEEE, 2007, pp. 1072--1079.

\bibitem{xliu2013}
X.~Liu, A.~Bouguettaya, J.~Wu, and L.~Zhou, ``Ev-lcs: A system for the
  evolution of long-term composed services,'' \emph{IEEE Transactions on
  Services Computing}, vol.~6, no.~1, pp. 102--115, 2013.

\bibitem{lieicsoc2017}
L.~Qu, A.~Bouguettaya, and A.~G. Neiat, ``Confidence-aware reputation
  bootstrapping in composite service environments,'' in \emph{Service-Oriented
  Computing}, 2017, pp. 158--174.

\bibitem{DBLP:conf/wecwis/NguyenYZ12}
H.~T. Nguyen, J.~Yang, and W.~Zhao, ``Bootstrapping trust and reputation for
  web services,'' in \emph{Proceedings of {CEC}}.\hskip 1em plus 0.5em minus
  0.4em\relax Hangzhou: IEEE, 2012, pp. 41--48.

\bibitem{DBLP:conf/wise/SkopikSD09}
F.~Skopik, D.~Schall, and S.~Dustdar, ``Start trusting strangers? bootstrapping
  and prediction of trust,'' in \emph{Proceedings of {WISE}}.\hskip 1em plus
  0.5em minus 0.4em\relax Heidelberg: Springer-Verlag Berlin, 2009, pp.
  275--289.

\bibitem{7401117}
Q.~Wu, F.~Ishikawa, Q.~Zhu, and D.~Shin, ``Qos-aware multigranularity service
  composition: Modeling and optimization,'' \emph{IEEE Transactions on Systems,
  Man, and Cybernetics: Systems}, vol.~46, no.~11, pp. 1565--1577, 2016.

\bibitem{DBLP:conf/atal/BurnettNS10}
C.~Burnett, T.~J. Norman, and K.~Sycara, ``Bootstrapping trust evaluations
  through stereotypes,'' in \emph{Proceedings of AAMAS}, 2010, pp. 241--248.

\bibitem{DBLP:journals/vldb/MalikB09}
Z.~Malik and A.~Bouguettaya, ``Rateweb: Reputation assessment for trust
  establishment among web services,'' \emph{{VLDB} Journal}, vol.~18, no.~4,
  pp. 885--911, 2009.

\bibitem{boutilier2004cp}
C.~Boutilier, R.~I. Brafman, C.~Domshlak, H.~H. Hoos, and D.~Poole, ``Cp-nets:
  A tool for representing and reasoning with conditional ceteris paribus
  preference statements,'' \emph{Journal of Artificial Intelligence Research},
  vol.~21, pp. 135--191, 2004.

\bibitem{dorigo2016ant}
L.~M. Gambardella and M.~Dorigo, ``Ant-q: A reinforcement learning approach to
  the traveling salesman problem,'' in \emph{Machine Learning Proceedings},
  1995, pp. 252 -- 260.

\bibitem{watkins1992q}
C.~J. Watkins and P.~Dayan, ``Q-learning,'' \emph{Machine learning}, vol.~8,
  no. 3-4, pp. 279--292, 1992.

\bibitem{Ricci2015}
F.~Ricci, L.~Rokach, and B.~Shapira, \emph{Recommender Systems: Introduction
  and Challenges}.\hskip 1em plus 0.5em minus 0.4em\relax Boston, MA: Springer
  US, 2015, pp. 1--34.

\bibitem{Kalloori:2017:ICS:3079628.3079696}
S.~Kalloori and F.~Ricci, ``Improving cold start recommendation by mapping
  feature-based preferences to item comparisons,'' in \emph{Proceedings of
  UMAP}, 2017, pp. 289--293.

\bibitem{XiongPeerTrust}
L.~Xiong and L.~Liu, ``Peertrust: supporting reputation-based trust for
  peer-to-peer electronic communities,'' \emph{IEEE Trans. on Knowledge and
  Data Engineering}, vol.~16, no.~7, pp. 843--857, 2004.

\bibitem{DBLP:conf/atal/HuynhJS06}
T.~D. Huynh, N.~R. Jennings, and N.~R. Shadbolt, ``Certified reputation: how an
  agent can trust a stranger,'' in \emph{Proceedings of {AAMAS}}, 2006, pp.
  1217--1224.

\bibitem{DBLP:journals/eis/WuZL15}
Q.~Wu, Q.~Zhu, and P.~Li, ``A neural network based reputation bootstrapping
  approach for service selection,'' \emph{Enterprise {(IS)}}, vol.~9, no.~7,
  pp. 768--784, 2015.

\bibitem{DBLP:conf/icwe/YahyaouiZ11}
H.~Yahyaoui and S.~Zhioua, ``Bootstrapping trust of web services through
  behavior observation,'' in \emph{Proceedings of the ICWE}.

\bibitem{DBLP:conf/icsoc/HuangLNFCT14}
K.~Huang, Y.~Liu, S.~Nepal, Y.~Fan, S.~Chen, and W.~Tan, ``A novel equitable
  trustworthy mechanism for service recommendation in the evolving service
  ecosystem,'' in \emph{Proceedings of the {ICSOC}}, 2014, pp. 510--517.

\bibitem{cornelio2013updates}
C.~Cornelio, J.~Goldsmith, N.~Mattei, F.~Rossi, and K.~B. Venable, ``Updates
  and uncertainty in cp-nets,'' in \emph{Proceedings of the AJCAI}, 2013, pp.
  301--312.

\bibitem{ramirez2017evolutionary}
A.~Ram{\'\i}rez, J.~A. Parejo, J.~R. Romero, S.~Segura, and A.~Ruiz-Cort{\'e}s,
  ``Evolutionary composition of qos-aware web services: a many-objective
  perspective,'' \emph{Expert Systems with Applications}, vol.~72, pp.
  357--370, 2017.

\bibitem{Borges:2016:PPG:2972958.2972966}
H.~Borges, A.~Hora, and M.~T. Valente, ``Predicting the popularity of github
  repositories,'' in \emph{Proceedings of PROMISE}, 2016, pp. 9:1--9:10.

\bibitem{7792740}
Q.~{Zhao}, C.~{Wang}, P.~{Wang}, M.~{Zhou}, and C.~{Jiang}, ``A novel method on
  information recommendation via hybrid similarity,'' \emph{IEEE Trans. on
  Systems, Man, and Cybernetics: Systems}, vol.~48, no.~3, pp. 448--459, 2018.

\bibitem{sajibicsoc2016}
S.~Mistry, A.~Bouguettaya, H.~Dong, and A.~Erradi, ``Qualitative economic model
  for long-term iaas composition,'' in \emph{Proceedings of ICSOC}, 2016, pp.
  317--332.

\bibitem{7060671}
P.~{Wang}, Z.~{Ding}, C.~{Jiang}, M.~{Zhou}, and Y.~{Zheng}, ``Automatic web
  service composition based on uncertainty execution effects,'' \emph{IEEE
  Trans. on Services Computing}, vol.~9, no.~4, pp. 551--565, 2016.

\bibitem{7027801}
Y.~{Wu}, C.~{Yan}, Z.~{Ding}, G.~{Liu}, P.~{Wang}, C.~{Jiang}, and M.~{Zhou},
  ``A multilevel index model to expedite web service discovery and composition
  in large-scale service repositories,'' \emph{IEEE Transactions on Services
  Computing}, vol.~9, no.~3, pp. 330--342, 2016.

\bibitem{wang2012wcp}
H.~Wang, J.~Zhang, W.~Sun, H.~Song, G.~Guo, and X.~Zhou, ``Wcp-nets: a weighted
  extension to cp-nets for web service selection,'' in \emph{Proceedings of
  ICSOC}, 2012, pp. 298--312.

\bibitem{Wang2016124}
H.~Wang, S.~Shao, X.~Zhou, C.~Wan, and A.~Bouguettaya, ``Preference
  recommendation for personalized search,'' \emph{Knowledge-Based Systems},
  vol. 100, no.~3, pp. 124 -- 136, 2016.

\bibitem{zhang2003time}
G.~P. Zhang, ``Time series forecasting using a hybrid arima and neural network
  model,'' \emph{Neurocomputing}, vol.~50, pp. 159--175, 2003.

\bibitem{Jian1531265}
D.~A. D'Mello and V.~Ananthanarayana, ``Dynamic selection mechanism for quality
  of service aware web services,'' vol.~4, no.~1.\hskip 1em plus 0.5em minus
  0.4em\relax Taylor \& Francis, 2010, pp. 23--60.

\bibitem{sutton1998reinforcement}
R.~S. Sutton, A.~G. Barto, F.~Bach \emph{et~al.}, \emph{Reinforcement learning:
  An introduction}.\hskip 1em plus 0.5em minus 0.4em\relax USA: MIT press,
  1998.

\bibitem{chen2013combinatorial}
W.~Chen, Y.~Wang, and Y.~Yuan, ``Combinatorial multi-armed bandit: General
  framework and applications,'' in \emph{Proceedings of JMLR}, 2013, pp.
  151--159.

\bibitem{bertsekas2018feature}
D.~P. Bertsekas, ``Feature-based aggregation and deep reinforcement learning: A
  survey and some new implementations,'' \emph{IEEE/CAA Journal of Automatica
  Sinica}, vol.~6, no.~1, pp. 1--31, 2018.

\bibitem{9049451}
M.~{Ghahramani}, Y.~{Qiao}, M.~C. {Zhou}, A.~{O'Hagan}, and J.~{Sweeney},
  ``Ai-based modeling and data-driven evaluation for smart manufacturing
  processes,'' \emph{IEEE/CAA Journal of Automatica Sinica}, vol.~7, no.~4, pp.
  1026--1037, 2020.

\bibitem{sajibbook2018}
S.~Mistry, A.~Bouguettaya, and H.~Dong, \emph{Long-Term Qualitative IaaS
  Composition}.\hskip 1em plus 0.5em minus 0.4em\relax Springer, 2018, pp.
  77--110.

\bibitem{8456329}
R.~{Xiong}, J.~{Wang}, Z.~{Li}, B.~{Li}, and P.~C.~K. {Hung}, ``Personalized
  lstm based matrix factorization for online qos prediction,'' in
  \emph{Proceedings of the ICWS}, 2018, pp. 34--41.

\bibitem{silver2016mastering}
D.~Silver, A.~Huang, C.~J. Maddison, A.~Guez, L.~Sifre, G.~Van Den~Driessche,
  J.~Schrittwieser, I.~Antonoglou, V.~Panneershelvam, M.~Lanctot \emph{et~al.},
  ``Mastering the game of go with deep neural networks and tree search,''
  \emph{nature}, vol. 529, no. 7587, p. 484, 2016.

\bibitem{beller2017travistorrent}
M.~Beller, G.~Gousios, and A.~Zaidman, ``Travistorrent: Synthesizing travis ci
  and github for full-stack research on continuous integration,'' in
  \emph{Proceedings of the ICMSR}, 2017, pp. 447--450.

\bibitem{githubDev}
\BIBentryALTinterwordspacing
G.~Developer, ``Github graphql api,'' 2017. [Online]. Available:
  \url{https://developer.github.com/v4/}
\BIBentrySTDinterwordspacing

\end{thebibliography}
%



%
\vspace{-15mm}
\begin{IEEEbiography}[{\includegraphics[width=1in,height=1.25in,clip,keepaspectratio]{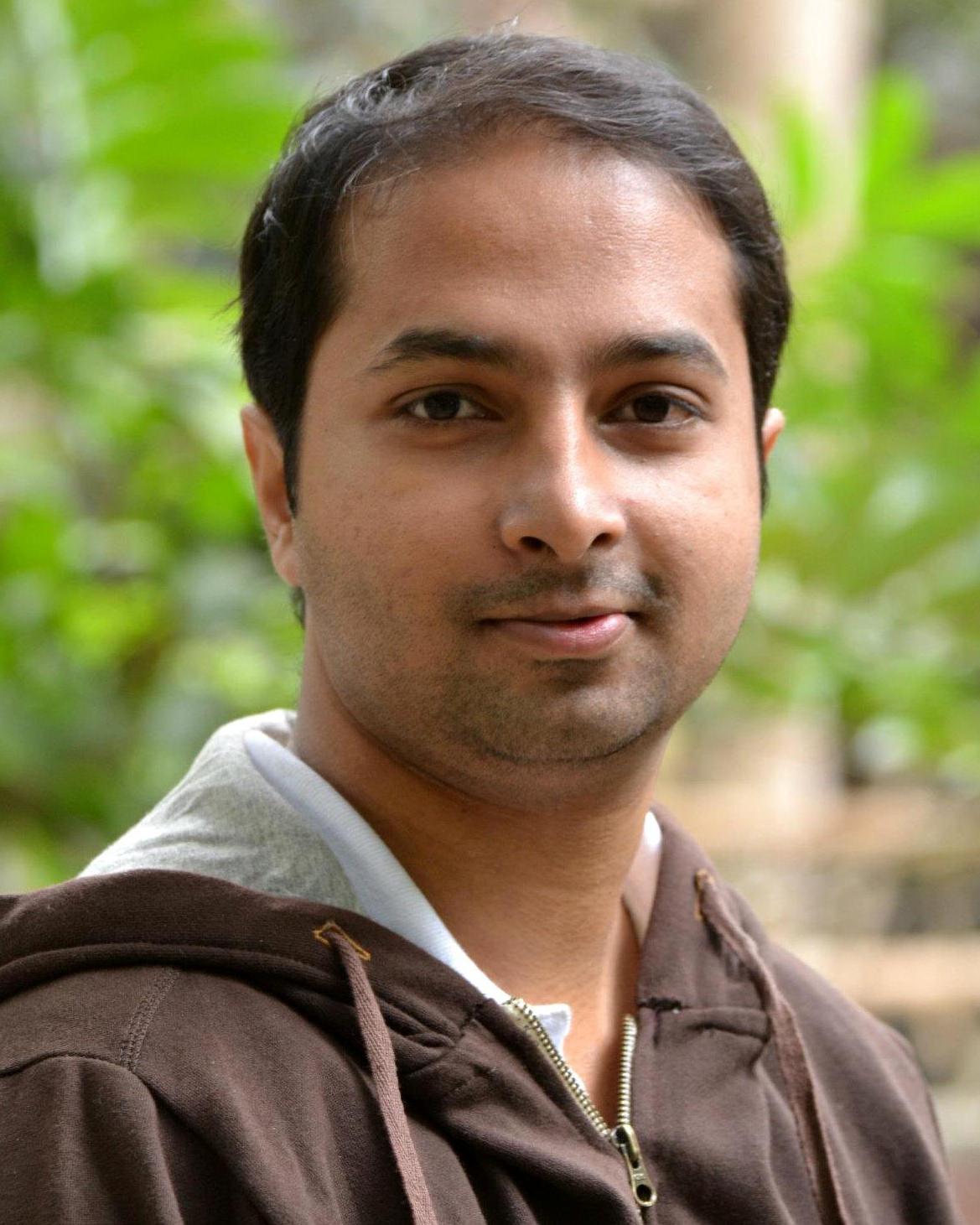}}]{Sajib Mistry} is lecturer at School of Elect Eng, Computer and Math Sci in Curtin University, Australia. He was previously a postdoc fellow at School of Computer Science in University of Sydney. He received his PhD from the RMIT University, Australia in 2017. His research interests include Edge/Cloud Computing, Big Data and IoT. He has published articles in international journals and conferences, such as IEEE TSC, TKDE, ACM CACM, ACM TOI, ACM TWEB, ICSOC, WISE, ICWS and SCC. He received the best paper award in ICSOC 2016. 
\end{IEEEbiography}
\vspace{-15mm}
\begin{IEEEbiography}[{\includegraphics[width=1in,height=1.25in,clip,keepaspectratio]{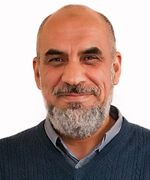}}]{Athman Bouguettaya}
is Professor in the School of Computer Science at the University of Sydney, Australia. He received his PhD in Computer Science from the University of Colorado at Boulder (USA) in 1992. He was previously Science Leader in Service Computing at CSIRO ICT Centre, Canberra. Australia. Before that, he was a tenured faculty member and Program director in the Computer Science department at Virginia Polytechnic Institute and State University (commonly known as Virginia Tech) (USA). He is or has been on the editorial boards of several journals including, the IEEE Transactions on Services Computing, ACM Transactions on Internet Technology, the International Journal on Next Generation Computing and VLDB Journal. He has published more than 250 books, book chapters, and articles in journals and conferences in the area of databases and service computing (e.g., the IEEE TKDE, the ACM TWEB, WWW Journal, VLDB Journal, SIGMOD, ICDE, VLDB, and EDBT). He was the recipient of several federally competitive grants in Australia (e.g., ARC) and the US (e.g., NSF, NIH). He is a Fellow of the IEEE and a Distinguished Scientist of the ACM.
\end{IEEEbiography}








\end{document}